\documentclass{article}
\usepackage{arxiv,times}
\usepackage{amsmath,amsfonts,bm}

\def\eqref#1{equation~\ref{#1}}
\def\1{\bm{1}}

\DeclareMathAlphabet{\mathsfit}{\encodingdefault}{\sfdefault}{m}{sl}
\SetMathAlphabet{\mathsfit}{bold}{\encodingdefault}{\sfdefault}{bx}{n}

\usepackage{amsmath,amssymb,booktabs,array}
\PassOptionsToPackage{dvipsnames}{xcolor}
\usepackage{graphicx,hyperref,url,xcolor,microtype,xspace,enumitem,placeins}
\DeclareRobustCommand{\doi}[1]{doi: \begingroup\urlstyle{rm}%
  \hypersetup{hidelinks}\href{https://doi.org/#1}{\nolinkurl{#1}}\endgroup}
\usepackage{tcolorbox}
\tcbuselibrary{breakable,skins}
\newtcolorbox[auto counter]{frozenprompt}[1]{enhanced,breakable,
  colback=cyan!5!white,colframe=cyan!50!black,
  colbacktitle=cyan!75!black,coltitle=white,
  title={Prompt~\thetcbcounter. #1},fonttitle=\footnotesize\bfseries,
  fontupper=\footnotesize\ttfamily,boxrule=0.6pt,arc=2pt,
  left=7pt,right=7pt,top=3pt,bottom=3pt,before skip=3pt,after skip=3pt}
\usepackage[font=footnotesize]{caption}
\usepackage{minitoc}

\renewcommand{\eqref}[1]{\textup{(\ref{#1})}}
\newcommand{\act}{\textsc{Act}\xspace}
\newcommand{\askpref}{\textsc{Ask-Preference}\xspace}
\newcommand{\askobj}{\textsc{Ask-Objective}\xspace}
\newcommand{\relaxaction}{\textsc{Relax}\xspace}
\newcommand{\ifc}{\ensuremath{\mathrm{IFC}}\xspace}
\newcommand{\oss}{\textsc{gpt-oss-120b}\xspace}
\newcommand{\astra}{GPT-6-Astra\xspace}
\newcommand{\ind}{\mathbf{1}}
\title{\parbox{\textwidth}{\centering\textcolor{purple}{Ask, Relax, or Act?\\Evaluating Actionable Indeterminacy\\in LLM Preference Reasoning}}}
\usepackage{authblk}

\author[1]{Ang Li}
\author[1]{Yue Lin}
\author[2]{Feifei Kou}
\author[3]{Zhan Su}
\author[3]{Prayag Tiwari}
\author[4]{\\Wenhao Li}
\author[5]{Shuhui Zhu}
\author[1]{Hongyuan Zha}
\author[1,*]{Baoxiang Wang}
\affil[1]{The Chinese University of Hong Kong Shenzhen}
\affil[2]{Beijing University of Posts and Telecommunications}
\affil[3]{Halmstad University College}
\affil[4]{Tongji University}
\affil[5]{University of Waterloo}
\hypersetup{colorlinks=true,allcolors=purple,pdfborder={0 0 0}}
\iclrfinalcopy
\begin{document}
\doparttoc
\faketableofcontents
\maketitle
\begingroup
\renewcommand{\thefootnote}{\fnsymbol{footnote}}
\footnotetext[1]{Corresponding author: \href{mailto:bxiangwang@gmail.com}{\texttt{bxiangwang@gmail.com}}.}
\endgroup

\begin{abstract}
An LLM agent can recognize uncertainty yet still choose the wrong next step:
asking when action is already justified, or seeking clarification when the
constraints must change.
We formalize \emph{actionable indeterminacy}: act when an accepted action is
shared across all admissible preferences or objectives, clarify when each
possibility is feasible but no action is shared, and propose a minimum-cost
permitted constraint repair when the request is infeasible.
We construct a solver-grounded benchmark spanning object allocation, meeting
scheduling, apartment choice, and stable matching.
Matched pairs retain the same source while changing whether intervention is
necessary, and evaluation separates decision correctness, matched-pair
reliability, and fully correct responses.
Our findings reveal a recurring difficulty in recognizing when intervention
is unnecessary: models can identify situations requiring clarification or
repair yet still intervene when a justified action already exists.
Correct decision labels also fail to guarantee usable actions, questions,
or repairs.
Crucially, response requirements shape not only how decisions are expressed
but also which decisions are made.
Making the required content explicit substantially improves fully correct
responses and can change intervention decisions, even when outputs are
already parseable.
These findings highlight that reliable agency requires more than recognizing
uncertainty: it requires intervening only when necessary and translating the
chosen next step into a verifiable response.
\end{abstract}

\section{Introduction}
Knowing that information is missing is not the same as knowing that it is
needed~\citep{zhang2025necessary}. An LLM agent may ask an unnecessary question when a justified action
already exists, act prematurely when an unresolved preference is pivotal, or
seek clarification when the request is actually infeasible. These failures
require different remedies. The central question is therefore \textit{not simply
whether uncertainty is present, but whether it prevents action and, if so,
what intervention can resolve the obstacle}.

Consider a group choosing an apartment with one unresolved preference
comparison. If the same apartment remains optimal under either ordering, the
agent need not ask. Likewise, an unspecified collective objective need not
block action when the admissible objectives share an optimal choice~\citep{lang2020collective}.
Clarification becomes necessary when each admissible case is feasible but no
action is accepted in all of them. By contrast, if no apartment satisfies the
budget and commute requirements, learning more about preferences cannot
restore feasibility: the agent must propose a permitted constraint change~\citep{junker2004quickxplain}.
The distinction is between uncertainty that can remain unresolved and an
obstacle that must be addressed before acting.

Evaluating this distinction requires more than recognizing an uncertainty
source or producing the right decision label. A model that asks whenever a
preference is missing may succeed when clarification is necessary yet fail
when it is not. A response can also select the correct intervention label
while omitting the pivotal question or failing to provide a valid repair
and resulting action. Evaluation must therefore separate whether
intervention is warranted, whether both sides of an action/intervention
contrast are handled correctly, and whether the response satisfies the
required content.

\begin{figure}[t]
\centering
\includegraphics[width=\linewidth]{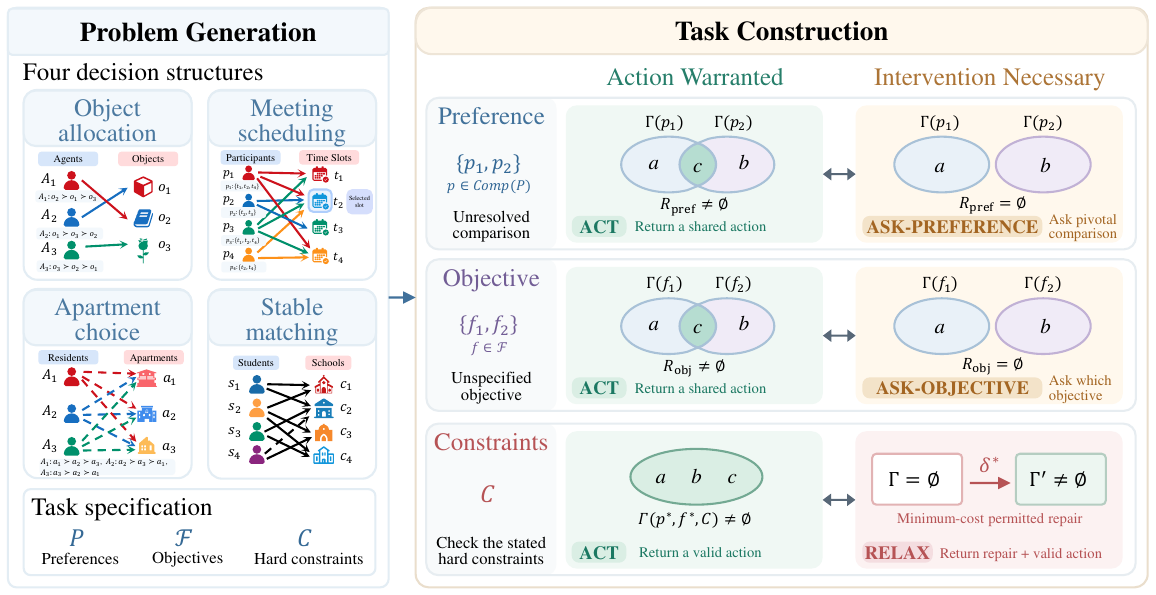}
\caption{Overview of the actionable-indeterminacy benchmark.
Four decision structures generate tasks specified by preferences $P$,
objectives $\mathcal F$, and constraints $C$. Each six-task group contains
three matched pairs. For preference and objective uncertainty, a shared
accepted action warrants \act; otherwise the model must ask a pivotal
preference or objective question. For constraints, a feasible request warrants
\act, while an infeasible request requires \relaxaction: a minimum-cost
permitted repair and a valid resulting action. Each pair retains the source
while changing whether intervention is necessary.}
\label{fig:framework}
\end{figure}

We formalize these distinctions as \emph{actionable indeterminacy} and
construct a solver-grounded benchmark spanning object allocation, meeting
scheduling, apartment choice, and stable matching. The model selects \act,
\askpref, \askobj, or \relaxaction and supplies the corresponding action,
question, or minimum-cost permitted repair with a valid resulting action.
Tasks vary one source at a time; each six-task group contains three matched
pairs that retain the source while changing whether intervention is
necessary. Exact solvers determine the warranted labels, and fixed
checkers evaluate the required answer content without an LLM judge.
We further control selected local cues and compare response requirements
on identical logical tasks to examine how these affect intervention
behavior. Figure~\ref{fig:framework} summarizes the benchmark and its
three action/intervention contrasts.

\subsection{Main Results}
\label{sec:main-findings}
Our main V3 evaluation spans \astra, Claude-5-Sonnet, Gemini-3.1-Pro,
and \oss. We distinguish decision Accuracy, Intervention Flip Correctness
(\ifc), which requires both decisions in a matched pair to be correct,
and Fully correct (FC), which additionally checks the required response
content. Our findings are as follows:

\begin{enumerate}[leftmargin=*,itemsep=5pt,topsep=4pt]
\item \textbf{Recognizing when not to intervene is a central difficulty.}
Under the original V3 prompts, most decision failures for \astra and
Claude-5-Sonnet occur on tasks where action is already warranted, despite
strong performance on necessary-intervention cases. Recognizing situations
that require clarification or repair therefore does not ensure that a model
can avoid unnecessary intervention. Matched-pair evaluation makes this
two-sided requirement explicit. The pair metric measures joint decision success
(Section~\ref{sec:main-results}).

\item \textbf{Response requirements shape decisions, not just their expression.}
Models make errors even on pairs with matched local features, yet making the
required response explicit substantially improves decision success on those
same tasks. This aggregate pattern appears in \astra, Claude-5-Sonnet,
and Gemini-3.1-Pro, although the affected sources and remaining difficulties
differ across models. Controlling selected local cues is therefore not
sufficient to characterize intervention errors independently of how the
required answer is specified (Section~\ref{sec:local-cues}).

\item \textbf{Correct labels do not establish verifiable responses.}
Models can select the warranted response category while failing the checks
for its action, question, or repair. In the same-task comparisons, explicit
content requirements largely close this gap, with further aggregate gains
when model-generated evidence is requested. For \astra, all responses are
already parseable in every condition, so the improvement cannot be
attributed to JSON syntax alone. Claude-5-Sonnet and Gemini-3.1-Pro exhibit
the same aggregate improvement pattern. The response checks capture both content and representation errors
(Section~\ref{sec:requirements}; Appendix~\ref{app:scoring}).

\end{enumerate}

\subsection{Related Work}
\label{sec:related-work}
\paragraph{Preference reasoning and collective decisions.}
Classical incomplete-preference reasoning distinguishes possible from
necessary outcomes
\citep{konczak2005voting,walsh2007uncertainty,xia2011winners,lang2020collective},
while vote elicitation and regret-based decisions connect information
acquisition to collective choices
\citep{lu2011robust,oren2013efficient,boutilier2016incomplete}.
Language-model research studies preference elicitation
\citep{handa2024bayesian,andukuri2024stargate}, collective representation
\citep{fish2023generative}, preference-constrained matching
\citep{fauchard2025reasoning}, and reasoning under indeterminacy
\citep{hosseini2026preference}. These foundations motivate evaluating the
decision relevance of unresolved information: an accepted action can remain
available even when preferences or objectives are not fully specified.

\paragraph{Uncertainty and selective clarification.}
Self-evaluation, semantic uncertainty, and linguistic calibration examine how
models assess and communicate answer reliability
\citep{kadavath2022know,kuhn2023semantic,band2024calibration}.
Clarification research addresses ambiguous information needs in search and
question answering
\citep{aliannejadi2019asking,min2020ambigqa,kuhn2022clam,zhang2024clamber},
while interactive training connects question quality to subsequent task
success \citep{andukuri2024stargate,zhang2025future}.
Decision-focused elicitation and clarification policies already assess
whether further interaction is worthwhile
\citep{boutilier2006minimax,zhang2025necessary,suri2025structured,ta2026regret}.
A shared-action criterion provides a complementary evaluation target:
intervention necessity follows from the task's admissible possibilities,
rather than the model's confidence or a general dialogue-cost objective.

\paragraph{Reasoning and response evaluation.}
Reasoning prompts and decomposition
\citep{wei2022chain,kojima2022zero,zhou2023least}, path aggregation and search
\citep{wang2023consistency,yao2023tree}, and environment interaction
\citep{yao2023react} shape how models solve tasks.
Prompt-sensitivity studies
\citep{lu2022ordered,webson2022prompts,sclar2023formatting} and fine-grained
instruction-following benchmarks
\citep{zhou2023ifeval,jiang2024followbench,qin2024infobench} further motivate
controlling presentation and checking the requested response content.
The resulting distinction is between selecting the warranted decision and
expressing an action, question, or repair that satisfies the evaluation
requirements.
An extended discussion appears in
Appendix~\ref{app:related-work}.

\section{Problem Formulation and Methodology}
\label{sec:formulation-setup}
\subsection{Problem Formulation}
\paragraph{Problem domains.}
We consider object allocation, meeting scheduling, apartment choice, and
stable matching. Let $Y$ be the finite action space, $P$ the observed
preferences, $\mathcal F$ the admissible objectives, and $C$ the hard
constraints. For complete preferences $p$ and objective $f$,
$\Gamma(p,f,C)\subseteq Y$ contains the actions accepted by the stated rule:
all tied optima, a specified algorithm's output, or all feasible actions,
depending on the task (Appendix~\ref{app:structure-rules}).

\paragraph{Actionability.}
Each task varies one source $s\in\{\mathrm{pref},\mathrm{obj},\mathrm{spec}\}$
(preferences, objectives, or constraints), keeping the remaining preferences
and objectives fixed at $p^\star$ and $f^\star$. Writing
$\operatorname{Comp}(P)$ for the complete rankings consistent with $P$, define
\begin{equation}
\mathcal R_{\mathrm{pref}}=\bigcap_{p\in\operatorname{Comp}(P)}
\Gamma(p,f^\star,C),\quad
\mathcal R_{\mathrm{obj}}=\bigcap_{f\in\mathcal F}\Gamma(p^\star,f,C),\quad
\mathcal R_{\mathrm{spec}}=\Gamma(p^\star,f^\star,C).
\label{eq:source-robust-sets}
\end{equation}
A nonempty $\mathcal R_s$ warrants \act with an action from that set;
the shared action need not be unique. If every admissible case is feasible
but the intersection is empty, \askpref or \askobj must ask the pivotal
preference or objective question. An infeasible constraint task instead
requires \relaxaction: a minimum-cost permitted repair and an action valid
under that same repair. These single-source rules do not prioritize
simultaneous sources. Formal decision and repair definitions appear in
Appendix~\ref{app:structure-rules}.

\subsection{Methodology and Experimental Setup}
\paragraph{Task construction.}
Each V3 group contains three source-matched action/intervention pairs,
yielding six independent tasks. Each pair changes whether intervention is
necessary. Preference tasks hide one adjacent comparison, giving two
admissible completions.
Four structures, three sizes ($n\in\{4,6,8\}$), and three generator seeds
form 36 cells; the common evaluation uses 20 complete groups per cell,
totaling 4,320 prompts. Exact solvers supply labels, and fixed checkers
assess responses without an LLM judge. Local-cue (H5) and output-requirement
(H2) comparisons use distinct sampling designs; construction and sampling
details are in Appendix~\ref{app:protocol-inventory}.

\paragraph{Models and evaluation procedure.}
We evaluate \astra~\citep{openai2026astra},
Claude-5-Sonnet~\citep{anthropic2026sonnet5},
Gemini-3.1-Pro~\citep{googledeepmind2026gemini31pro}, and
\oss~\citep{openai2025gptoss} on V3; the first three also cover H2 and H5.
Single-turn requests use fresh conversations without tools or access to
private labels; task-completion follow-ups retain their conversation history.
Within-model comparisons are primary because inference settings and
collection conditions differ. Configuration, record checks, and recovery
procedures are documented in Appendix~\ref{app:reproducibility}.

\paragraph{Evaluation metrics.}
Accuracy measures decision-label correctness. Intervention Flip Correctness
(\ifc) is the fraction of matched pairs with both labels correct, aggregated
by pair counts. Fully correct (FC) additionally requires accepted response
content; all-six accuracy requires all six labels in a group to be correct.
FC measures acceptance under the fixed interface and checker;
evidence validity is scored separately
(Appendix~\ref{app:scoring}). Unparseable completed responses and unresolved
OSS truncations remain unsuccessful, with the latter treated as delivery
failures rather than observed reasoning errors (Appendix~\ref{app:oss-v3}).

\section{When Is Intervention Necessary?}
\label{sec:main-results}
We examine whether models distinguish a visible source of uncertainty from a
necessary intervention, then ask whether their decisions translate into
accepted responses and successful task completion.

\subsection{Individual Decisions and Matched-Pair Reliability}
Strong individual decision accuracy coexists with failures to handle both
sides of a matched action/intervention contrast. On the common 4,320-prompt
V3 subset, \astra achieves 92.57\% Accuracy and 85.56\% IFC, while
all six decisions are correct in 65.00\% of groups. These measures distinguish
getting an individual response category right from correctly handling every
source-matched contrast in a group. Claude-5-Sonnet, Gemini-3.1-Pro, and \oss
show the same distinction between individual, pair, and group success;
complete results and seed-wise breakdowns are reported in
Appendix~\ref{app:v3}. OSS scores retain unresolved truncations in the full
denominator, with complete-response results and missing-response bounds
reported separately in Appendix~\ref{app:oss-v3}.

Astra's observed IFC is close to its descriptive independence reference of
85.28\%, calculated from the action- and intervention-member accuracies
within each cell and source (Appendix~\ref{app:independence}). Matched-pair evaluation measures joint success on both sides
of the actionability distinction.

\begin{figure}[t]
\centering
\captionsetup{skip=0pt}
\includegraphics[width=\linewidth]{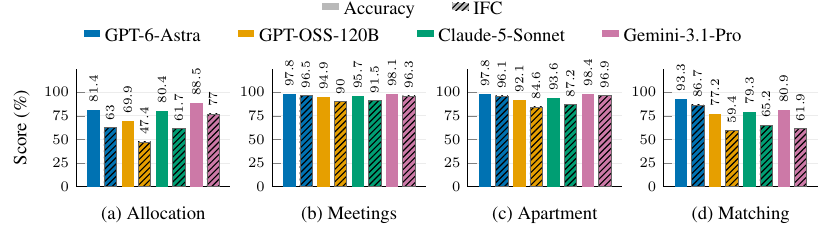}%
\caption{V3 results by structure on the identical 4,320-task subset:
1,080 tasks and 540 pairs per structure and model. GPT-OSS-120B retains truncated responses
in the full denominator, as in Figure~\ref{fig:v3-source}.
Solid bars show Accuracy and hatched bars show IFC;
colors identify models. Values above bars are percentages.}
\label{fig:v3}
\end{figure}

\subsection{Errors Depend on Structure and Source}
The dominant decision difficulty for Astra and Claude under the original V3
prompts is recognizing when intervention is \emph{unnecessary}. For Astra,
303 of 321 incorrect or unparseable decisions occur on tasks where action is
warranted; for Claude, the corresponding count is 468 of 551 decision errors.
Both models correctly identify most constructed intervention-needed cases,
yet repeatedly fail to act when the available information already suffices.
Success at recognizing the need for clarification or repair therefore does
not ensure that a model avoids unnecessary intervention.

This asymmetry is accompanied by substantial variation across structures and
sources. Across the evaluated configurations, meeting scheduling and apartment
choice are handled more reliably than object allocation and stable matching
(Figure~\ref{fig:v3}). The source profiles are not identical, however:
preference pairs have the highest IFC for Astra and Gemini, whereas constraint
pairs have the highest IFC for Claude and OSS. Objective pairs are especially
difficult for Claude and OSS, and Gemini also makes unnecessary interventions
on many objective-source action tasks (Figure~\ref{fig:v3-source}). The source ordering varies across models. OSS uses a distinct inference
implementation and budget and retains incomplete responses. Section~\ref{sec:local-cues}
examines whether selected local cues help explain the observed errors.

\begin{figure}[t]
\centering
\captionsetup{skip=0pt}
\includegraphics[width=\linewidth]{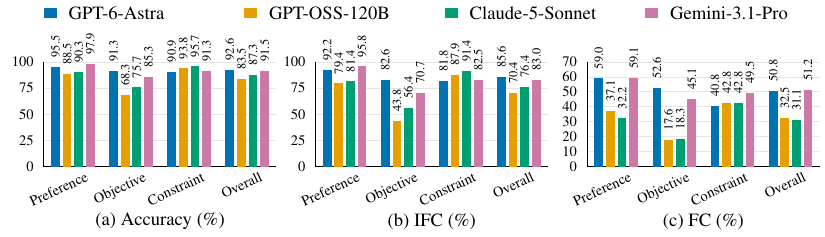}
\caption{V3 results on the identical 4,320-task subset. Each source contains
1,440 tasks and 720 pairs; Overall aggregates all three sources.
Colors identify models as in Figure~\ref{fig:v3}. Values above bars are
percentages rounded to one decimal place; higher is better.
FC uses the original strict response checks. OSS retains 92 truncated responses as unsuccessful;
its scores use the full denominator (Appendix~\ref{app:oss-v3}).
Exact values appear in Table~\ref{tab:v3-source}; Gemini collection and
scoring details are in Appendix~\ref{app:gemini-v3}.}
\label{fig:v3-source}
\end{figure}

\subsection{Correct Labels Do Not Establish Executable Answers}
Under the original V3 checks, correct decision labels frequently accompany
responses that fail the required content checks. This separation appears in
all four models (Figure~\ref{fig:v3-source}). For Astra, FC is 50.79\%, despite
its high decision Accuracy. The clearest example is relaxation: all 720
\relaxaction labels are correct, yet none of the responses passes the complete
repair-and-action check. Every response omits or nulls the required top-level
\texttt{action}, so no six-task group passes all response-content checks.
Some inspected answers place a resulting action in another field, while the
original wording does not consistently specify the exact representation
required by the checker. This identifies a mismatch between requested and
checked content.

The other models reinforce the distinction without sharing an identical
failure profile. Claude's rejected responses mix unrecognized field names or
action representations with actions outside the accepted set, even though all
responses are parseable JSON (Appendix~\ref{app:claude-v3-errors}). Gemini
likewise has many correct decisions whose content is rejected
(Appendix~\ref{app:gemini-v3}). For OSS, even the missing-response upper bound
on FC remains far below its observed Accuracy, so unresolved truncation alone
cannot explain the separation (Appendix~\ref{app:oss-v3}). The response checks are detailed in Appendix~\ref{app:scoring}.
Section~\ref{sec:requirements} compares output requirements on the same tasks.

\subsection{Task Completion after Intervention}
\label{sec:completion}
A warranted intervention is an intermediate step, not a completed task.
Building on evaluation of clarification through the usefulness of its
answers~\citep{rao2018questions}, we test whether models can proceed from an
accepted intervention to a correct final action. We select 540
intervention-needed tasks by retaining three members from each of five
matched groups in every cell. Astra, Claude-5-Sonnet, and Gemini-3.1-Pro use
the same tasks and explicit JSON requirements. In this two-step procedure,
the first response must specify the intervention and leave \texttt{action}
null until feedback. A valid first intervention then receives feedback and a
final-action request in the same conversation; an invalid first response
receives no second turn.
The independent reference route receives predetermined feedback directly
(Appendix~\ref{app:completion}).

The results separate failures to initiate the required interaction from
failures to solve the updated task. Astra completes 99.07\% of tasks
autonomously, close to the reference rate of 99.81\%; its few failures include
both premature actions and incorrect final actions after valid interventions.
For Claude, most autonomous failures occur before an accepted intervention
has been produced, often because an action is supplied before feedback.
Additional failures follow valid interventions, so first-step compliance and
final-action correctness must be assessed separately. Detailed rates and
recovery accounting, including the two authorized Claude restarts, are
reported in Appendix~\ref{app:completion}.

Gemini illustrates an even stronger dependence on the response requirements:
501 of 540 first responses wrap JSON in Markdown code fences and are rejected
by the fixed bare-JSON parser, preventing feedback and a second turn.
Code-fence rejection also affects its autonomous final and reference
responses. Completion rates include these parser rejections. The comparison uses
intervention-needed tasks; the two routes differ in conversation history
and can use different permitted repairs. Appendix~\ref{app:completion}
reports first-step validity and final-action success separately.

\section{Do Local Cues Explain the Errors?}
\label{sec:local-cues}
We ask whether the errors observed in Section~\ref{sec:main-results} persist
when selected local cues are matched, and whether the answer depends on how
the required response is specified.

\subsection{Local-Cue-Matched Pairs}
A source-matched pair can still differ in local features that correlate with
its labels. Motivated by documented prompt
sensitivity~\citep{sclar2023formatting,chatterjee2024posix}, the H5 collection
matches selected source-dependent features while changing whether intervention
is necessary. For example, objective pairs preserve the objective menu and
per-objective solution counts; allocation-constraint pairs preserve agent and
object degrees in permission graphs while differing in whether a complete
assignment exists. A deterministic rule using only these identical features
cannot distinguish the two members. Solution-count matching is enforced during construction;
the solutions remain private. Appendix~\ref{app:local-design} details the
feature checks and construction differences from V3.

We select five of the ten available pairs per source and cell using a fixed
seed without consulting model outcomes, yielding 540 pairs and 1,080 logical
tasks across all 36 cells. These are two-task pairs, not complete six-task
groups. Each task is answered under its original requirements and with
appended explicit JSON requirements, identical for both members of a pair.
The four requests per pair are randomized and use fresh conversations.
Astra's two conditions are collected within the same run; Claude-5-Sonnet
follows the same prompts and order, and a separate Gemini-3.1-Pro replication
uses the same fixed subset and four-call design. Each model contributes
2,160 responses. Claude's quota-related time gaps and Gemini's recovery
batches are documented in Appendix~\ref{app:local-design}.

\subsection{Matched Cues Do Not Make Errors Format-Invariant}
Errors remain on locally matched tasks, but their frequency changes markedly
with response requirements. For \astra, IFC rises from 76.67\% to 98.33\%
when the required response is made explicit, alongside an increase in
individual Accuracy (Figure~\ref{fig:local-cue}). Same-task corrections greatly
outnumber reversals. The improvement concerns decision-label correctness,
not only whether the accompanying response content passes the FC checks.

\begin{figure}[t]
\centering
\captionsetup{skip=0pt}
\includegraphics[width=\linewidth]{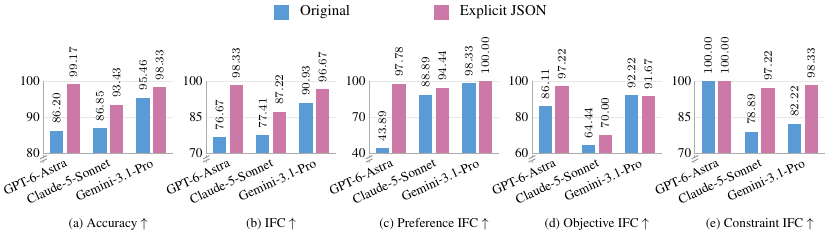}
\caption{Local-cue comparison under original and explicit JSON requirements.
Each model uses the same 1,080 tasks and 540 pairs per condition;
each source contributes 180 pairs. Blue bars denote the original
requirements and pink bars denote explicit JSON. Labels show percentages;
higher is better in all panels. Vertical axes are truncated, as marked:
the ranges in (a)--(e) are 80--100\%, 70--100\%, 40--100\%,
60--100\%, and 70--100\%, respectively. Bar heights are not comparable across panels.
Exact values are retained in Tables~\ref{tab:h5} and~\ref{tab:gemini-h5}; strict FC results are
reported in Appendix~\ref{app:local-design}.}
\label{fig:local-cue}
\end{figure}

The aggregate decision improvement replicates in Claude-5-Sonnet and
Gemini-3.1-Pro, but its source profile differs. Astra's largest IFC gain is on
preference pairs, while its constraint decisions are already perfect in both
conditions. Claude and Gemini benefit most on constraint pairs. Objective
pairs remain the least reliable source for Claude under explicit requirements,
and Gemini's objective IFC decreases slightly despite its aggregate gain.
Response specification affects decision success across the three models.
Strict FC also improves; detailed results
are reported in Appendix~\ref{app:local-design}, with Gemini's replication
separately documented in Appendix~\ref{app:gemini-h5}.

Errors persist with matched local features and vary with response
requirements. H5 uses a separately constructed task collection, including
a different allocation-constraint family. Section~\ref{sec:requirements}
compares three output requirements on the same V3 logical instances,
measuring decision and response-content correctness separately.

\section{How Much Do Output Requirements Matter?}
\label{sec:requirements}
We now hold the logical tasks fixed to distinguish improvements in required
response content from changes in decision correctness. We also examine whether
supplying candidate answers or emphasizing intervention availability produces
similar benefits.

\subsection{Three Requirements for the Same Tasks}
Following the distinction between content and format
constraints~\citep{jiang2024followbench}, H2 compares three response
requirements on the same 2,160 logical tasks and 1,080 pairs, drawn from ten
complete V3 groups per cell. Each task is answered under its original prompt,
explicit JSON requirements, and the same JSON requirements plus
model-generated evidence. \astra, Claude-5-Sonnet, and Gemini-3.1-Pro each
contribute 6,480 responses. The original condition retains any existing
request for structured output; it does not mean that JSON was never requested.

The explicit condition specifies allowed decision strings, field names,
action representations, and source-specific question and correction
structures. In particular, a relaxation must include both a permitted change
and a valid resulting action in the required top-level field. The evidence
condition additionally requests witnesses derived from the public task,
such as actions under each preference completion or objective, or an account
of infeasibility. It supplies no gold actions, labels, or solved instances.
The public task facts and core scoring rules remain unchanged across H2
conditions, while evidence is checked separately and does not enter FC
(Appendix~\ref{app:output}). These comparisons therefore vary the
\emph{specification of the response}, not just its punctuation: field
descriptions and evidence requests can guide the content of the solution.

\begin{table}[t]
\centering\small
\caption{Output-requirement comparison (\%). For each model and condition,
Accuracy and FC use 2,160 tasks, while IFC uses 1,080 pairs.
Evidence is checked separately from FC. Correct counts are retained
in Tables~\ref{tab:h2-counts} and~\ref{tab:gemini-h2-counts} in Appendix~\ref{app:output}.}
\label{tab:h2}
\setlength{\tabcolsep}{1.5pt}
\begin{tabular}{@{}l*{9}{>{\centering\arraybackslash}p{34pt}}@{}}
\toprule
& \multicolumn{3}{c}{GPT-6-Astra}
& \multicolumn{3}{c}{Claude-5-Sonnet}
& \multicolumn{3}{c}{Gemini-3.1-Pro}\\
\cmidrule(lr){2-4}\cmidrule(lr){5-7}\cmidrule(lr){8-10}
Requirements & Accuracy & IFC & FC & Accuracy & IFC & FC & Accuracy & IFC & FC\\
\midrule
Original & 92.59 & 85.74 & 50.79 & 85.74 & 74.07 & 32.18 & 92.08 & 84.35 & 50.83\\
Explicit JSON & 94.58 & 89.17 & 94.40 & 90.56 & 81.67 & 87.13 & 95.79 & 91.57 & 95.79\\
JSON + evidence & 97.59 & 95.19 & 97.45 & 94.21 & 88.70 & 91.81 & 96.99 & 93.98 & 96.99\\
\bottomrule
\end{tabular}
\end{table}

\subsection{Most Execution Failures Disappear with Explicit Requirements}
Making the required content explicit largely closes the gap between correct
decisions and checker-accepted responses (Table~\ref{tab:h2}). For \astra,
FC rises from 50.79\% to 94.40\% with explicit requirements, reaching 97.45\%
when evidence is additionally requested. Claude and Gemini show the same
aggregate pattern. Correct-label responses that fail the content checks
become rare for Astra, disappear for Gemini in both explicit conditions, and
decrease sharply for Claude. Thus, the original separation between Accuracy
and FC is highly sensitive to how the required response is described.

All Astra
responses are already parseable in every condition, and parsing failures are
rare for Claude and Gemini. A parseable response can still omit a required
action, underspecify a clarification question, or fail to express an accepted
repair. Explicit requirements address this broader mismatch between choosing
the warranted category and supplying its required content. H2 applies the same core acceptance checks in every condition
(Appendix~\ref{app:scoring}).

\subsection{Judgment Also Changes, but Not Equally Across Sources}
The benefit extends beyond expressing a decision that was already correct.
Across all three models, both Accuracy and IFC improve with explicit
requirements and improve further in aggregate when evidence is requested
(Table~\ref{tab:h2}). Response specification therefore affects which next
step is selected, not only whether the accompanying content is accepted.
The evidence condition has the strongest aggregate results. Witness validity
is assessed separately in Appendix~\ref{app:output}.

The effects vary by source.
Explicit JSON improves Astra's preference decisions but lowers its constraint
Accuracy; adding evidence then improves constraint decisions while slightly
reducing preference Accuracy. Claude's constraint Accuracy instead increases
with explicit JSON and then decreases slightly with evidence. The direction of these source-level changes differs across models. The earlier OSS comparison also
reports a different decision trend on its separate collection
(Appendix~\ref{app:legacy-format}). Response requirements affect intervention decisions, with source-dependent
gains and reversals.

\subsection{Separating Answer Generation from Intervention Reminders}
We use two complementary comparisons to examine answer construction and
intervention emphasis. Both use the same 1,080 tasks for Astra, Claude, and
Gemini: five complete six-task groups per cell, sampled without consulting
model answers or scores. The selection comparison presents four candidate
responses, including exactly one fully correct candidate, and uses each
model's existing same-task H2 explicit-JSON generation responses as its
baseline. The wording comparison instead collects new responses under neutral
wording and an intervention reminder, retaining identical task facts and
explicit JSON requirements. Appendix~\ref{app:selection-framing} provides the
construction, record checks, and detailed results.

Supplying candidate answers does not uniformly improve intervention decisions
(Table~\ref{tab:selection-main}). Astra and Gemini improve in aggregate,
whereas Claude deteriorates. All three models improve on constraint FC, but
this common benefit does not extend uniformly to the other sources. For
Claude, preference and objective performance decline, and unnecessary
interventions nearly double. The selection condition changes answer construction and supplies both
correct answer content and distractors.

Intervention reminders yield only modest incremental gains once the response
requirements are already explicit (Figure~\ref{fig:framing-main}). Unnecessary
interventions decrease slightly for all three models, while missed
interventions decrease for Claude and remain unchanged for Astra and Gemini.
Paired comparisons contain both improvements and reversals. These effects
measure the addition of an intervention reminder to explicit content requirements. Taken together, the comparisons show that
specifying response content, providing candidate answers, and emphasizing
intervention availability produce distinct behavioral effects rather than
interchangeable improvements.

\begin{table}[htbp]
\centering\small
\caption{Generating versus selecting an answer.
All entries are percentages. For each model and condition, Accuracy and
FC use 1,080 matched tasks, while IFC uses 540 pairs. Correct counts are
retained in Appendix~\ref{app:selection-framing}.
The generation baseline uses only the same selected tasks from H2.}
\label{tab:selection-main}
\setlength{\tabcolsep}{3pt}
\begin{tabular}{@{}lccccccccc@{}}
\toprule
& \multicolumn{3}{c}{GPT-6-Astra}
& \multicolumn{3}{c}{Claude-5-Sonnet}
& \multicolumn{3}{c}{Gemini-3.1-Pro}\\
\cmidrule(lr){2-4}\cmidrule(lr){5-7}\cmidrule(lr){8-10}
Response mode & Acc. & IFC & FC & Acc. & IFC & FC & Acc. & IFC & FC\\
\midrule
Generate (explicit JSON) & 94.72 & 89.44 & 94.63 & 90.56 & 81.67 & 87.50 & 95.56 & 91.11 & 95.56\\
Select (four options) & 97.59 & 95.19 & 97.59 & 83.43 & 67.04 & 83.15 & 96.76 & 93.52 & 96.76\\
\bottomrule
\end{tabular}
\end{table}

\begin{figure}[htbp]
\centering
\captionsetup{skip=0pt}
\includegraphics[width=\linewidth]{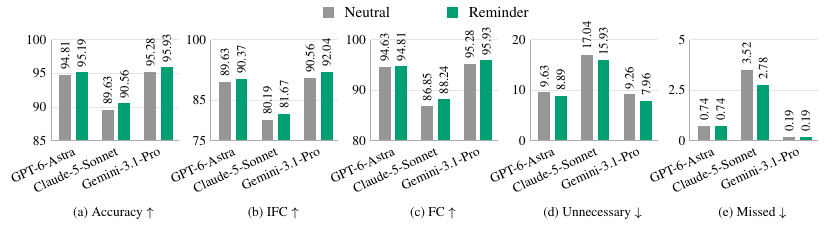}
\par
\caption{Effect of an intervention reminder on the same 1,080 tasks
and 540 pairs per model. Gray bars denote neutral
wording and teal-green bars denote the reminder. Labels show percentages,
calculated from original counts. Panels (a)--(c) show performance
(higher is better); panels (d)--(e) show intervention errors (lower is better).
Exact scores and counts are in Table~\ref{tab:framing-detail}.}
\label{fig:framing-main}
\end{figure}

\section{Conclusion}
We introduced \emph{actionable indeterminacy} to distinguish uncertainty that
can remain unresolved from obstacles that require clarification or constraint
repair before action. Our solver-grounded benchmark makes this distinction
testable through matched action/intervention pairs across four decision
structures. Under the original prompts, models can recognize necessary
interventions yet still intervene when action is already justified.
Crucially, response requirements shape both the decisions models make and
whether their accompanying content is accepted. Making the required content
explicit substantially improves fully correct responses and can change
intervention decisions even when outputs are already parseable, although
gains vary across models and sources. These findings support evaluating decision correctness, matched-pair
reliability, and response-content validity separately.
The central lesson is that reliable agency requires more than detecting
uncertainty: it requires recognizing what must be clarified or repaired,
what can remain unresolved, and how to express the warranted next step in
a verifiable form.

\section*{AI Use Statement}
We used generative AI tools for language polishing and code assistance. The authors take full responsibility for the final content of this work.

\bibliographystyle{arxiv}
\bibliography{references}
\clearpage
\appendix
\part{Appendix}
\parttoc
\clearpage
\section{Extended Related Work}
\label{app:related-work}
\paragraph{Preference reasoning and decision-focused elicitation.}
Classical work on incomplete preferences distinguishes outcomes that can win
under some completion from those that win under every completion
\citep{pini2007incompleteness,betzler2010dichotomy,xia2011winners}. Utility, vote, and matching elicitation
connect information acquisition to the decision at hand, including whether
the information already obtained suffices to stop asking
\citep{ha1997problem,chajewska2000adaptive,boutilier2002pomdp,conitzer2002elicitation,kalech2011practical,drummond2014preference}. Regret-based methods
further seek useful decisions under imprecise utilities with limited user
interaction \citep{wang2003incremental,boutilier2006minimax,braziunas2007minimax}. Language-model research extends
preference elicitation and collective representation to natural-language
settings, and evaluates reasoning under epistemic and structural
indeterminacy \citep{handa2024bayesian,fish2023generative,hosseini2026preference}.
Our shared-action criterion draws on this decision-focused tradition: an
action can remain accepted across all admissible possibilities even when
information is incomplete. A nonempty intersection requires neither a unique
outcome nor identical solution sets. This exact acceptance criterion differs
from a general minimax-regret policy, which can select a compromise with
positive worst-case regret. We evaluate whether models recognize this
boundary through matched action/intervention pairs.

\paragraph{Selective clarification and information value.}
Clarification research studies the recovery of user information needs and
question generation \citep{aliannejadi2019asking,kumar2020clarq,aliannejadi2020clariq},
alongside ambiguity detection and resolution
\citep{xu2019clarification,min2020ambigqa,lee2023clarification,zhang2024clamber}.
Question ranking and generation model answer utility
\citep{rao2018questions,rao2019answer}; selective clarification and interactive
training connect asking to task success
\citep{hu2020interactive,khalid2020combining,kuhn2022clam,andukuri2024stargate,zhang2025future}.
Methods also use predictive uncertainty in grounded dialogue
\citep{naszadi2023aligning}, uncertainty over user intents and tool arguments,
and interaction costs \citep{zhang2025necessary,suri2025structured}; interactive benchmarks
evaluate missing-intent clarification, false-premise correction, and
multi-turn stopping policies \citep{zhao2026askbench,ta2026regret}.
These studies already treat clarification as a decision problem. Our
complementary emphasis is a controlled finite setting that retains the
uncertainty source within each pair while changing whether a shared accepted
action exists. Exact solvers provide the labels, making unnecessary and
missed interventions separately observable. This tests intervention necessity
under specified rules without optimizing a general dialogue-cost objective.

\paragraph{Infeasibility and constraint repair.}
Over-constrained reasoning distinguishes explaining why requirements conflict
from finding changes that restore feasibility
\citep{reiter1987diagnosis,liffiton2008algorithms,liffiton2016enumeration,previti2018premise}.
Preferred explanations and
relaxations can account for priorities among constraints
\citep{junker2004quickxplain,felfernig2012diagnosis}.
Partial and valued constraint frameworks also formalize violation costs
\citep{freuder1992partial,schiex1995valued}.
This provides a separate foundation for the \relaxaction branch: clarification
of a preference or objective does not itself repair an infeasible request.
Our tasks specify permitted edits and a cost function, and require a
minimum-cost repair with a valid resulting action. This requirement differs
from finding an inclusion-minimal conflict or a preference-maximal consistent
set; these notions need not select the same repair. We incorporate minimum-cost repair into a common intervention evaluation.

\paragraph{Confidence, ambiguity recognition, and intervention.}
Self-evaluation and calibration study answer reliability
\citep{jiang2021calibration,kadavath2022know,yin2023know,chen2023calibration};
semantic uncertainty measures uncertainty over meanings
\citep{kuhn2023semantic,farquhar2024entropy}, while linguistic calibration
examines communicated confidence
\citep{tian2023calibration,krause2023confidently,band2024calibration}.
Ambiguity-sensitive evaluations test recognition of multiple interpretations
\citep{liu2023ambiguity}; behavioral studies distinguish recognition from
actually asking a clarifying question \citep{su2026knowing}.
These distinctions are relevant, but recognizing ambiguity alone does not
establish that intervention is necessary. Our labels depend on the task's
information and accepted actions: a confident model can choose an action
that fails under an admissible possibility, while a model uncertain about
its own computation may face a task that already permits justified action.
The benchmark therefore evaluates a task-level intervention policy without
treating confidence estimates as ground truth or IFC as a calibration metric.

\paragraph{Reasoning procedures and verifiable response evaluation.}
Reasoning traces and planning \citep{wei2022chain,wang2023plansolve},
executable reasoning \citep{gao2023pal,chen2023programthoughts},
and aggregation, feedback, and environmental interaction
\citep{wang2023consistency,yao2023react,shinn2023reflexion,madaan2023selfrefine}
offer complementary routes to improved task performance.
Prompt-sensitivity studies \citep{sclar2023formatting,zhu2024promptrobust,chatterjee2024posix}
and instruction-following benchmarks \citep{zhou2023ifeval,wen2024complexbench}
motivate evaluating how tasks and response requirements are expressed.
Our evaluations check the warranted response and its required content:
Accuracy measures decision correctness, IFC measures both decisions in a
matched pair, and FC additionally checks answer content. Explicit requirements
can change both answer expression and the selected decision, so their effects
cannot be reduced to JSON parseability. Answer selection and intervention
reminders further probe behavior. Reading these comparisons together helps distinguish decision
errors from failures to express a checkable action, question, or repair;
Appendix~\ref{app:scoring} documents the scope of the fixed checkers.

\section{V3 Construction and Complete Results}
\label{app:v3}
The full V3 collection contains 36 cells, 40 complete six-task groups per
cell, and 8,640 tasks. Astra and Claude use the identical stratified subset of 20 groups per cell:
2,160 \act tasks and 720 tasks for each of \askpref, \askobj, and
\relaxaction. Every structure contributes 1,080 tasks; every size and
generator seed contributes 1,440 tasks. Sampling complete groups preserves the
three source contrasts. The seed identifiers are 20260901, 20260902, and
20260903. They control generation, not repeated decoding of the same prompt.

In allocation, an action maps agents to distinct objects. In scheduling and
apartment choice, an action selects an available slot or apartment under the
stated collective rule and constraints. In stable matching, an action maps
the two sides of the market, with the admissible rule specified in each task.
The preference branch of stable matching uses the A-proposing deferred
acceptance outcome under each completion.
The finite generators and structure-specific solution rules determine
the accepted actions.

% Begin inlined appendix: apartment_worked_example.tex
\clearpage
\subsection{A Worked Example: Six Matched Apartment-Choice Tasks}
\label{app:apartment-example}
This example is a synthetic benchmark group with fixed identifier
\texttt{v3::apartment\_choice::n4::s20260901::set003}; it also appears in
the H2 source collection. We abbreviate \texttt{Apartment1}--\texttt{Apartment4}
as $A_1$--$A_4$ and the four residents as $R_1$--$R_4$.
The six tasks are independent requests, not successive turns. We show gold
decisions and oracle calculations, not a model's observed responses.

\begin{center}
\captionof{table}{The fixed apartment attributes and complete preference profile.
Rankings are provided directly; they are not inferred from the attributes.}
\label{tab:apartment-example-input}
\small
\begin{tabular}{lrrr@{\qquad}ll}
\toprule
Apartment & Rent & Commute (min) & Area (m$^2$) & Resident & Ranking\\
\midrule
$A_1$ & 9,500 & 36 & 63 & $R_1$ & $A_4>A_3>A_1>A_2$\\
$A_2$ & 10,000 & 39 & 77 & $R_2$ & $A_4>A_2>A_1>A_3$\\
$A_3$ & 10,000 & 18 & 55 & $R_3$ & $A_1>A_3>A_2>A_4$\\
$A_4$ & 7,000 & 52 & 69 & $R_4$ & $A_2>A_4>A_1>A_3$\\
\bottomrule
\end{tabular}
\end{center}

\paragraph{Preference pair: the location of missing information matters.}
Both tasks fix the egalitarian objective: minimize the worst resident rank
of the selected apartment (rank 1 is best). Under the complete profile, the
four apartments have worst ranks $(3,4,4,4)$, so $A_1$ is uniquely optimal.
The first task hides the order of $A_3$ and $A_1$ for $R_1$:
\[
R_1:\quad A_4>\{A_3,A_1\}_{\text{unknown order}}>A_2.
\]
If $A_3>A_1$, the worst-rank vector is $(3,4,4,4)$; if $A_1>A_3$,
it is still $(3,4,4,4)$. Thus the two optimal sets intersect at $\{A_1\}$.
The gold decision is \act with $A_1$, despite the missing preference.

The second task instead hides $R_1$'s order of $A_1$ and $A_2$:
\[
R_1:\quad A_4>A_3>\{A_1,A_2\}_{\text{unknown order}}.
\]
The completion $A_1>A_2$ gives worst ranks $(3,4,4,4)$ and optimum $A_1$;
the completion $A_2>A_1$ gives $(4,3,4,4)$ and optimum $A_2$.
Their intersection is empty. The gold decision is \askpref, asking $R_1$
to compare $A_1$ and $A_2$. Both tasks omit one adjacent comparison from the
same resident, but only the second omission prevents a robust choice.

\paragraph{Objective pair: an unspecified rule need not block action.}
These tasks restore all complete rankings. Utilitarian aggregation minimizes
the sum of resident ranks; rank-maximal aggregation lexicographically maximizes
the counts of first-place, then second-place, and subsequent ranks;
egalitarian aggregation minimizes the worst rank.
\begin{center}
\small
\begin{tabular}{lrrrr}
\toprule
Quantity & $A_1$ & $A_2$ & $A_3$ & $A_4$\\
\midrule
Sum of ranks & 10 & 10 & 12 & 8\\
First-place votes & 1 & 1 & 0 & 2\\
Worst rank & 3 & 4 & 4 & 4\\
\bottomrule
\end{tabular}
\end{center}
If the admissible rules are utilitarian and rank-maximal, both select $A_4$,
so the gold decision is \act with $A_4$. If egalitarian is also admissible,
its optimum is $A_1$ and the three-way intersection is empty. The gold
decision is \askobj; the oracle clarification distinguishes utilitarian
from egalitarian aggregation. Two rules supporting $A_4$ do not justify a
majority vote over rules: no such meta-rule is specified in the task.

\clearpage
\paragraph{Constraint pair: infeasibility calls for repair, not clarification.}
Both tasks use the complete profile and the egalitarian rule, now restricted
to apartments satisfying the hard requirements. With rent at most 10,000
and commute at most 18 minutes, only $A_3$ is feasible. Consequently the
gold decision is \act with $A_3$, even though $A_1$ was optimal without
these constraints.

The paired task tightens the rent limit to 7,000 while retaining the
18-minute commute limit. $A_4$ satisfies the rent limit but violates the
commute limit; $A_3$ satisfies the commute limit but violates the rent limit;
$A_1$ and $A_2$ violate both. No apartment is feasible, so the gold decision
is \relaxaction. The exact certificate lists two permitted minimal repairs:
\begin{enumerate}[nosep]
\item Raise the rent limit from 7,000 to 10,000, leaving commute unchanged;
return $A_3$.
\item Raise the commute limit from 18 to 52 minutes, leaving rent unchanged;
return $A_4$.
\end{enumerate}
Each is the smallest sufficient threshold increase when modifying that
requirement alone. Minimality is relative to the benchmark's allowed edits;
it does not equate monetary costs with minutes of travel. The archived oracle
response uses the first repair; both appear in its certificate.

\begin{center}
\captionof{table}{Six independent gold decisions in the worked example.
Each adjacent pair varies whether intervention is necessary within one source.}
\label{tab:apartment-example-six}
\small
\begin{tabular}{@{}lp{0.49\linewidth}l@{}}
\toprule
Source & Task variation & Gold decision\\
\midrule
Preference & $R_1$'s $A_3/A_1$ order unknown & \act: $A_1$\\
& $R_1$'s $A_1/A_2$ order unknown & \askpref\\
\addlinespace
Objective & Utilitarian or rank-maximal & \act: $A_4$\\
& Egalitarian also admissible & \askobj\\
\addlinespace
Constraint & Rent $\leq10{,}000$, commute $\leq18$ & \act: $A_3$\\
& Rent $\leq7{,}000$, commute $\leq18$ & \relaxaction\\
\bottomrule
\end{tabular}
\end{center}

\paragraph{Connection to the evaluation.}
IFC requires both decision labels within a source pair to be correct; six-task
decision correctness requires all three pairs to be correct. FC additionally
checks the requested action, clarification, or repair and resulting action.
Full correctness requires both the correct label and accepted response content.
The crucial distinction is not whether missing information or constraints
are present, but whether they prevent a justified action.

This example has singleton optimal sets. The general criterion in
Eq.~\eqref{eq:source-robust-sets} is broader: \act requires a nonempty
intersection, not identical optimal sets, a unique shared action, or agreement
among merely a majority of admissible possibilities.
\clearpage
% End inlined appendix: apartment_worked_example.tex

\subsection{V3 Results and Gemini Replication}
\label{app:gemini-v3}

\begin{table}[h]
\centering\small
\caption{V3 results on the identical 4,320-task subset (\%). Each source contains
1,440 tasks and 720 pairs. FC is measured under the original strict
response checks.
OSS retains 92 truncated responses as unsuccessful; its columns are
full-denominator delivery scores (Appendix~\ref{app:oss-v3}).}
\label{tab:v3-source}
\setlength{\tabcolsep}{3pt}
\resizebox{\linewidth}{!}{%
\begin{tabular}{lrrrrrrrrrrrr}
\toprule
& \multicolumn{3}{c}{GPT-6-Astra}
& \multicolumn{3}{c}{GPT-OSS-120B}
& \multicolumn{3}{c}{Claude-5-Sonnet}
& \multicolumn{3}{c}{Gemini-3.1-Pro}\\
\cmidrule(lr){2-4}\cmidrule(lr){5-7}\cmidrule(lr){8-10}\cmidrule(lr){11-13}
Source & Accuracy & IFC & FC & Accuracy & IFC & FC & Accuracy & IFC & FC & Accuracy & IFC & FC\\
\midrule
Preference & 95.49 & 92.22 & 59.03 & 88.54 & 79.44 & 37.15 & 90.35 & 81.39 & 32.22 & 97.92 & 95.83 & 59.10\\
Objective & 91.32 & 82.64 & 52.57 & 68.26 & 43.75 & 17.64 & 75.69 & 56.39 & 18.33 & 85.35 & 70.69 & 45.07\\
Constraint & 90.90 & 81.81 & 40.76 & 93.82 & 87.92 & 42.78 & 95.69 & 91.39 & 42.78 & 91.25 & 82.50 & 49.51\\
\midrule
Overall & 92.57 & 85.56 & 50.79 & 83.54 & 70.37 & 32.52 & 87.25 & 76.39 & 31.11 & 91.50 & 83.01 & 51.23\\
\bottomrule
\end{tabular}
}
\end{table}

\paragraph{Gemini V3 replication.}
All 4,320 responses on the common subset are available, completing 2,160
pairs and 720 six-task groups. Accuracy is 3,953/4,320 (91.50\%), IFC is
1,793/2,160 (83.01\%), and FC is 2,213/4,320 (51.23\%). All six decisions
are correct in 437/720 groups (60.69\%), whereas all six answers pass FC
in 6/720 groups (0.83\%). Unnecessary intervention occurs on 244/2,160
action tasks (11.30\%); missed intervention occurs on 119/2,160
intervention-needed tasks (5.51\%).

\begin{table}[htbp]
\centering\small
\caption{Gemini V3 results by structure and size (\%): alternative breakdowns
of the same 4,320 tasks. Each structure has 1,080 tasks and 540 pairs;
each size has 1,440 tasks and 720 pairs.}
\label{tab:gemini-v3-strata}
\begin{tabular}{lrrr}
\toprule
Subset & Accuracy & IFC & FC\\
\midrule
Object allocation & 88.52 & 77.04 & 36.30\\
Meeting scheduling & 98.15 & 96.30 & 48.06\\
Apartment choice & 98.43 & 96.85 & 46.76\\
Stable matching & 80.93 & 61.85 & 73.80\\
\midrule
$n=4$ & 90.28 & 80.56 & 49.38\\
$n=6$ & 91.67 & 83.33 & 51.60\\
$n=8$ & 92.57 & 85.14 & 52.71\\
\bottomrule
\end{tabular}
\end{table}

The collection combines 3,118 inherited AGY responses, 176 initial API
responses, and two subsequent API batches of 513 responses each. Assignment
follows prior completion rather than randomization: 1,022 of the 1,026
batch responses are stable-matching tasks. Combined metrics retain all
309 pairs with responses from different batches and 174 six-task groups with responses from different batches;
differences between collection batches cannot isolate an access-interface effect. The AGY records lack
an independently verifiable served-model version; API responses report
\texttt{gemini-3.1-pro-preview}.

All responses parse under the fixed V3 parser, which accepts code fences.
Of the 3,953 responses scored as correct decisions, 1,740 fail FC.
Examples include alternative question-field names and textual repair
descriptions that are not accepted by the content checks; these failures
do not by themselves establish semantic incorrectness. The original scoring wrapper discarded label credit for 19 correctly
labelled objective-question responses after a content-check exception.
The reported metrics separate label scoring from content validation:
these responses receive decision credit but no FC credit, in accordance
with the Accuracy definition.
Package hashes, selected-response hashes, prompt hashes, unique IDs, and
group counts were checked, and all 4,320 original evaluations were reproduced with
the matching fixed verifier before the label-scoring correction. Original
scores are retained in the verification records; FC checks are unchanged. The original AGY response-validation records
are retained; the subsequent verification does not repeat every content check used when collecting responses.

\begin{table}[h]
\centering\small
\caption{Astra V3 label and response-content counts. A correct decision does not
require that the proposed action or question pass the strict content check.}
\label{tab:v3-labels}
\begin{tabular}{lrrr}
\toprule
Gold response & Tasks & Correct decisions & FC\\
\midrule
\act & 2,160 & 1,857 & 1,834\\
\askpref & 720 & 704 & 181\\
\askobj & 720 & 718 & 179\\
\relaxaction & 720 & 720 & 0\\
\midrule
Total & 4,320 & 3,999 & 2,194\\
\bottomrule
\end{tabular}
\end{table}
\begin{table}[h]
\centering\small
\caption{V3 results by size on the common subset. Each row has 1,440 tasks,
720 pairs, and 240 six-task groups. Values are percentages. OSS retains
0, 15, and 77 truncated responses at $n=4,6,8$, respectively, as unsuccessful.}
\begin{tabular}{llrrrr}
\toprule
Model & Size & Accuracy & IFC & FC & All-six accuracy\\
\midrule
GPT-6-Astra & $n=4$ & 92.15 & 84.86 & 50.69 & 60.42\\
& $n=6$ & 92.78 & 86.11 & 50.76 & 65.42\\
& $n=8$ & 92.78 & 85.69 & 50.90 & 69.17\\
\midrule
GPT-OSS-120B & $n=4$ & 84.31 & 70.97 & 32.57 & 29.17\\
& $n=6$ & 85.49 & 73.33 & 35.07 & 40.83\\
& $n=8$ & 80.83 & 66.81 & 29.93 & 35.42\\
\midrule
Claude-5-Sonnet & $n=4$ & 88.12 & 78.06 & 32.92 & 42.92\\
& $n=6$ & 88.33 & 78.33 & 32.22 & 48.33\\
& $n=8$ & 85.28 & 72.78 & 28.19 & 42.08\\
\bottomrule
\end{tabular}
\end{table}
Astra's IFC is nonmonotonic across the three sizes, whereas Accuracy is
nondecreasing and FC and all-six accuracy increase in this collection.
Complete cell-level counts are given in Table~\ref{tab:v3-cells}.

\subsection{Results across Generator Seeds}
\label{app:v3-seeds}
Table~\ref{tab:v3-seeds} separates the common 4,320-task subset by generator
seed. Each model is evaluated on the same 1,440 tasks and 720 pairs per seed.
Each seed contributes 120 tasks in each of the 12 structure--size cells,
with 40 tasks per source within each cell. Thus the composition and weights
are identical across seeds and models. We reuse the archived task-level
scores and require both members of each pair for IFC.

\begin{table}[h]
\centering\small
\caption{V3 results by generator seed on the common subset (\%). Each row
contains 1,440 tasks and 720 pairs. Incomplete OSS responses remain in the
denominator as unsuccessful deliveries; the last column reports their count.}
\label{tab:v3-seeds}
\begin{tabular}{llrrrr}
\toprule
Model & Generator seed & Accuracy & IFC & FC & Incomplete\\
\midrule
GPT-6-Astra & 20260901 & 92.85 & 85.97 & 51.04 & 0\\
& 20260902 & 93.19 & 86.67 & 51.11 & 0\\
& 20260903 & 91.67 & 84.03 & 50.21 & 0\\
\midrule
GPT-OSS-120B & 20260901 & 82.71 & 69.58 & 31.74 & 34\\
& 20260902 & 83.96 & 70.56 & 32.92 & 27\\
& 20260903 & 83.96 & 70.97 & 32.92 & 31\\
\midrule
Claude-5-Sonnet & 20260901 & 87.50 & 76.94 & 31.11 & 0\\
& 20260902 & 86.18 & 74.72 & 30.07 & 0\\
& 20260903 & 88.06 & 77.50 & 32.15 & 0\\
\bottomrule
\end{tabular}
\end{table}

Accuracy ranges from 91.67--93.19\% for Astra, 82.71--83.96\% for OSS,
and 86.18--88.06\% for Claude; IFC ranges from 84.03--86.67\%,
69.58--70.97\%, and 74.72--77.50\%, respectively.
The descriptive ordering Astra $>$ Claude $>$ OSS holds for both metrics
at every seed. Strict FC is lower than Accuracy for every model and seed;
its ordering is Astra $>$ OSS $>$ Claude throughout. These checks support
the stability of these aggregate comparisons across the three generated
collections. Seeds vary the generated tasks; each prompt has one selected response.
OSS differences also include variation in incomplete deliveries.
\subsection{OSS V3: Full Collection and Missing Responses}
\label{app:oss-v3}
The full OSS collection covers all 8,640 intended IDs, with 240 tasks in
each of 36 cells and no duplicate selected IDs. There are 8,455 complete
responses and 185 unresolved token-limit truncations. We retain the latter
in the denominator as unsuccessful deliveries, without calling them observed
reasoning errors or interpreting placeholder text as an answer.
The common 4,320-task subset used in the main text contains 4,228 complete
responses and 92 truncations. It is a subset of this collection, not a
separate run or an additional 4,320 observations.

Let $c_i$ indicate a complete response. Full-denominator Accuracy uses
$c_i d_i$, while its optimistic missing-response bound replaces that value
by $c_i d_i+(1-c_i)$. For FC, replace $d_i$ by $d_iv_i$.
Pair and six-task bounds take products of these terms within each unit.
An observed wrong answer therefore prevents that unit from succeeding even
when all its missing members are assumed correct. Complete-response scores
use only tasks with $c_i=1$; conditional pair and group scores require
all members to be complete. These bounds describe missing outcomes, not
sampling uncertainty or confidence intervals.

\begin{table}[htbp]
\centering\small
\caption{OSS V3 on all 8,640 tasks. Each entry is correct/total (\%).
The full-denominator column treats truncations as unsuccessful. The upper
bound treats only missing answers as correct; observed errors remain errors.}
\label{tab:oss-v3-full}
\setlength{\tabcolsep}{4pt}
\begin{tabular}{lrrr}
\toprule
Metric & Full denominator & Complete units only & Upper bound\\
\midrule
Accuracy & 7,211/8,640 (83.46) & 7,211/8,455 (85.29) & 7,396/8,640 (85.60)\\
IFC & 3,029/4,320 (70.12) & 3,029/4,152 (72.95) & 3,173/4,320 (73.45)\\
FC & 2,814/8,640 (32.57) & 2,814/8,455 (33.28) & 2,999/8,640 (34.71)\\
All-six accuracy & 502/1,440 (34.86) & 502/1,296 (38.73) & 526/1,440 (36.53)\\
All-six FC & 0/1,440 (0.00) & 0/1,296 (0.00) & 0/1,440 (0.00)\\
\bottomrule
\end{tabular}
\end{table}

Missing responses affect 168 pairs and 144 six-task groups. The individual
Accuracy bound spans 2.14 percentage points; the IFC bound spans 3.33 points,
so the task-level missing fraction must not be applied directly to pair
metrics. Even under the optimistic bound, FC is at most 34.71\%.
All-six FC remains zero because every group already contains an observed
failure. Thus truncation alone cannot account for the difference between
correct labels and fully correct responses under these checks.

\begin{table}[htbp]
\centering\small
\caption{Unresolved OSS truncations in the full collection. Each block
partitions the same 185 records; counts across blocks must not be added.}
\label{tab:oss-v3-missing}
\begin{tabular}{llrrr}
\toprule
Grouping & Condition & Tasks & Truncated & Rate (\%)\\
\midrule
Structure & Allocation & 2,160 & 115 & 5.32\\
& Meetings & 2,160 & 5 & 0.23\\
& Apartments & 2,160 & 9 & 0.42\\
& Matching & 2,160 & 56 & 2.59\\
\midrule
Size & $n=4$ & 2,880 & 1 & 0.03\\
& $n=6$ & 2,880 & 35 & 1.22\\
& $n=8$ & 2,880 & 149 & 5.17\\
\midrule
Source & Preference & 2,880 & 99 & 3.44\\
& Objective & 2,880 & 38 & 1.32\\
& Constraint & 2,880 & 48 & 1.67\\
\bottomrule
\end{tabular}
\end{table}

The missingness is concentrated rather than uniform: 149 of 185 truncations
occur at $n=8$, and 115 in allocation. The complete-response subset should
therefore not be interpreted as an unbiased replacement for the full collection.
On the common 4,320 tasks, complete-unit Accuracy, IFC, and FC are 85.36\%,
73.22\%, and 33.23\%; full-denominator values are 83.54\%, 70.37\%, and
32.52\%, with upper bounds 85.67\%, 73.61\%, and 34.65\%.

\paragraph{Generation and scoring checks.}
Complete responses from earlier batches were preserved. Selected records
comprise 3,609 attempts with a 2,048-token limit, 3,393 with a 3,072-token
limit, and 1,638 with a 6,144-token limit, including the remaining
truncations. No answer was chosen by correctness. Of the final 577 recovery
attempts, 515 completed and 62 truncated; together with 123 previously
preserved truncations, these give 185. The results describe this staged
recovery procedure, not a single fixed-budget generation run.

All 8,640 oracle positive controls passed, and all 8,455 complete responses
were scored again with identical results. Three complete responses failed
JSON parsing. One stable-matching response supplied objects where the
\texttt{choose\_between} field required options that could be compared as a
set. The original checker raised a type error; an explicit malformed-field
check retains decision-label credit but rejects that intervention and FC.
It does not repair the answer or suppress unrelated checker errors.
Replaying the existing 4,320 Astra and 4,320 Claude records through this
scoring function reproduced every stored evaluation. Original prompts,
answers, failed attempts, and file hashes remain archived.

\section{Independence Reference and Error Counts}
\label{app:independence}
Let $g$ index a cell--source group, $m_g$ its number of pairs, and
$p_{A,g}$ and $p_{I,g}$ the empirical decision accuracies on its action and
intervention members. The descriptive independence reference is
\begin{equation}
\mathrm{IFC}_{\mathrm{ind}}=
\frac{\sum_g m_g p_{A,g}p_{I,g}}{\sum_g m_g}.
\end{equation}
For Astra V3,
the observed value is 85.56\%, the reference is 85.28\%, and the difference
is +0.28 percentage points. The reference is computed descriptively from the same observed responses.
For Claude V3, observed IFC is 76.39\%, versus a 76.48\% independence
reference (a difference of $-0.09$ percentage points).

The archived mutually exclusive V3 response categories are: FC
(2,194), wrong decision (309), wrong clarification source (11), parse failure
(one), invalid action after a correct label (23), invalid preference question
(523), invalid objective question (539), and invalid relaxation/action content
(720). Invalid denotes rejection by the fixed checker. In particular, the question and relaxation
categories are affected by the original output wording.

\subsection{Claude V3 Response Content and H2 Comparison}
\label{app:claude-v3-errors}
Claude's mutually exclusive outcome counts are 1,344 FC responses,
551 incorrect decision labels, and 2,425 correct labels with rejected
content: 700 action answers, 525 preference questions, 530 objective
questions, and 670 relaxation answers. There are no parsing failures.
These categories follow the response checks in Appendix~\ref{app:scoring}.

\begin{table}[h]
\centering\small
\caption{Claude V3 label and response-content counts.}
\begin{tabular}{lrrr}
\toprule
Gold response & Tasks & Correct decisions & FC\\
\midrule
\act & 2,160 & 1,692 & 992\\
\askpref & 720 & 701 & 176\\
\askobj & 720 & 666 & 136\\
\relaxaction & 720 & 710 & 40\\
\midrule
Total & 4,320 & 3,769 & 1,344\\
\bottomrule
\end{tabular}
\end{table}

No response is repaired for the official score.

Claude H2 provides the relevant same-task output comparison: FC increases
from 695/2,160 (32.18\%) to 1,882/2,160 (87.13\%) and
1,983/2,160 (91.81\%). From original requirements to explicit JSON,
1,236 responses change from non-FC to FC and 49 change in the reverse
direction (646 remain FC and 229 remain non-FC). Adding evidence yields
161 improvements and 60 reversals (1,822 remain FC and 117 remain non-FC).
H2 uses a different subset from V3 and a broader acceptance rule for
minimum stable-matching repairs. Output effects are measured within
H2's matched comparison.

\begin{table}[h]
\centering\small
\caption{Claude V3 by structure. Each row contains 1,080 tasks and 540 pairs. Values are percentages.}
\begin{tabular}{lrrr}
\toprule
Structure & Accuracy & IFC & FC\\
\midrule
Apartments & 93.61 & 87.22 & 15.28\\
Meetings & 95.74 & 91.48 & 16.57\\
Allocation & 80.37 & 61.67 & 26.20\\
Matching & 79.26 & 65.19 & 66.39\\
\bottomrule
\end{tabular}
\end{table}
\begin{table}[p]
\centering\small
\caption{Claude-5-Sonnet V3 results across all 36 cells
(four structures $\times$ three sizes $\times$ three generator seeds), showing
variation across task strata and the gap between individual and joint success.
Entries are correct counts, with denominators in the headings: Acc. counts
correct decisions; IFC counts pairs with both decisions correct; FC additionally
requires valid response content. All-six and All-six FC count groups whose six
decisions or six full responses, respectively, are all correct. Seeds vary the
generated tasks, not repeated decoding of identical prompts. Low FC also reflects
strict response requirements.}
\label{tab:claude-v3-cells}
\setlength{\tabcolsep}{3pt}
\begin{tabular}{lrrrrrrr}
\toprule
Structure & $n$ & Seed & Acc./120 & IFC/60 & FC/120 & All-six/20 & All-six FC/20\\
\midrule
Apartments & 4 & 20260901 & 108 & 48 & 21 & 8 & 0\\
Apartments & 4 & 20260902 & 112 & 52 & 20 & 12 & 0\\
Apartments & 4 & 20260903 & 109 & 49 & 18 & 9 & 0\\
Apartments & 6 & 20260901 & 111 & 51 & 20 & 12 & 0\\
Apartments & 6 & 20260902 & 115 & 55 & 18 & 15 & 0\\
Apartments & 6 & 20260903 & 115 & 55 & 19 & 15 & 0\\
Apartments & 8 & 20260901 & 115 & 55 & 15 & 16 & 0\\
Apartments & 8 & 20260902 & 113 & 53 & 17 & 13 & 0\\
Apartments & 8 & 20260903 & 113 & 53 & 17 & 13 & 0\\
Meetings & 4 & 20260901 & 115 & 55 & 19 & 16 & 0\\
Meetings & 4 & 20260902 & 112 & 52 & 18 & 12 & 0\\
Meetings & 4 & 20260903 & 113 & 53 & 19 & 13 & 0\\
Meetings & 6 & 20260901 & 116 & 56 & 18 & 16 & 0\\
Meetings & 6 & 20260902 & 112 & 52 & 14 & 12 & 0\\
Meetings & 6 & 20260903 & 118 & 58 & 24 & 18 & 0\\
Meetings & 8 & 20260901 & 118 & 58 & 21 & 18 & 0\\
Meetings & 8 & 20260902 & 115 & 55 & 21 & 15 & 0\\
Meetings & 8 & 20260903 & 115 & 55 & 25 & 16 & 0\\
Allocation & 4 & 20260901 & 105 & 46 & 41 & 9 & 0\\
Allocation & 4 & 20260902 & 97 & 37 & 34 & 4 & 0\\
Allocation & 4 & 20260903 & 108 & 49 & 41 & 10 & 0\\
Allocation & 6 & 20260901 & 99 & 39 & 37 & 8 & 0\\
Allocation & 6 & 20260902 & 98 & 39 & 34 & 3 & 0\\
Allocation & 6 & 20260903 & 99 & 39 & 33 & 7 & 0\\
Allocation & 8 & 20260901 & 86 & 27 & 19 & 2 & 0\\
Allocation & 8 & 20260902 & 87 & 28 & 19 & 1 & 0\\
Allocation & 8 & 20260903 & 89 & 29 & 25 & 1 & 0\\
Matching & 4 & 20260901 & 94 & 40 & 77 & 2 & 0\\
Matching & 4 & 20260902 & 96 & 40 & 83 & 4 & 1\\
Matching & 4 & 20260903 & 100 & 41 & 83 & 4 & 1\\
Matching & 6 & 20260901 & 98 & 40 & 83 & 4 & 1\\
Matching & 6 & 20260902 & 93 & 40 & 81 & 2 & 0\\
Matching & 6 & 20260903 & 98 & 40 & 83 & 4 & 1\\
Matching & 8 & 20260901 & 95 & 39 & 77 & 1 & 0\\
Matching & 8 & 20260902 & 91 & 35 & 74 & 3 & 0\\
Matching & 8 & 20260903 & 91 & 37 & 76 & 2 & 1\\
\bottomrule
\end{tabular}
\end{table}

\section{Local-Cue Construction and Additional Results}
\label{app:local-design}
The H5 collection has 1,080 pairs, ten per source in each of 36 cells.
Feature equality is checked before model evaluation. Preference pairs match
the unresolved relation, the affected local preference list, agent identifiers,
the specified objective, and the counts of accepted actions in the two
completions. Objective pairs match the objective menu and description,
participant and candidate identifiers, and per-objective solution counts.
Constraint pairs use structure-specific feature checks. In allocation,
the permission graphs match the degree of each agent and object while differing
in the existence of a complete assignment. These graphs use permission
addition as the allowed repair, unlike the hard-requirement removal in V3
allocation. The comparison matches these selected features. Solution-count matching
is enforced during construction, with solutions kept private.

The same-run comparison samples five pairs per source and cell using
seed 20260910 without consulting response scores. It shuffles the 540 pair
blocks and randomly orders their four calls. The original and explicit
conditions receive the same logical tasks, with only the public output
requirements appended in the latter. The appended requirements are identical
for both members of a pair. The comparison uses previously evaluated tasks, with five pairs per source
and cell; results are aggregated by source and model.

\begin{table}[h]
\centering\small
\caption{Local-cue comparison (\%). Each reported model uses the same 1,080
tasks and 540 pairs per condition; each source contributes 180 pairs.
Main measures are Accuracy and IFC. Strict FC results are
supplementary.}
\label{tab:h5}
\setlength{\tabcolsep}{3pt}
\begin{tabular}{ll*{5}{>{\raggedleft\arraybackslash}p{45.5pt}}}
\toprule
Model & Requirements & Accuracy & IFC & Pref. IFC & Obj. IFC & Constr. IFC\\
\midrule
GPT-6-Astra & Original & 86.20 & 76.67 & 43.89 & 86.11 & 100.00\\
& Explicit JSON & 99.17 & 98.33 & 97.78 & 97.22 & 100.00\\
\midrule
Claude-5-Sonnet & Original & 86.85 & 77.41 & 88.89 & 64.44 & 78.89\\
& Explicit JSON & 93.43 & 87.22 & 94.44 & 70.00 & 97.22\\
\bottomrule
\end{tabular}
\end{table}

\begin{table}[h]
\centering\small
\caption{Astra same-run H5 results, with counts. Each source has 360 tasks and
180 pairs per condition; overall there are 1,080 tasks and 540 pairs.}
\label{tab:h5-detail}
\begin{tabular}{llrrr}
\toprule
Requirements & Source & Accuracy & IFC & FC\\
\midrule
Original & Preference & 65.56 (236) & 43.89 (79) & 34.17 (123)\\
 & Objective & 93.06 (335) & 86.11 (155) & 55.00 (198)\\
 & Constraint & 100.00 (360) & 100.00 (180) & 62.50 (225)\\
 & Overall & 86.20 (931) & 76.67 (414) & 50.56 (546)\\
\midrule
Explicit JSON & Preference & 98.89 (356) & 97.78 (176) & 98.89 (356)\\
 & Objective & 98.61 (355) & 97.22 (175) & 98.61 (355)\\
 & Constraint & 100.00 (360) & 100.00 (180) & 92.22 (332)\\
 & Overall & 99.17 (1,071) & 98.33 (531) & 96.57 (1,043)\\
\bottomrule
\end{tabular}
\end{table}
The same-task decision transitions are 929 correct-to-correct, 142
incorrect-to-correct, two correct-to-incorrect, and seven incorrect-to-incorrect.
For pair correctness, the counts are 413, 118, one, and eight, respectively.
One original-condition response is unparseable; none is unparseable in
the explicit condition. Stable-matching relaxation limitations apply to
FC in both conditions (Appendix~\ref{app:scoring}).

\begin{table}[htbp]
\centering\small
\caption{Claude-5-Sonnet H5 results. Percentages and
correct counts; each source has 360 tasks and 180 pairs per condition,
and each overall row has 1,080 tasks and 540 pairs.}
\label{tab:h5-claude-detail}
\begin{tabular}{llrrr}
\toprule
Requirements & Source & Accuracy & IFC & FC\\
\midrule
Original & Preference & 94.17 (339) & 88.89 (160) & 34.17 (123)\\
 & Objective & 80.56 (290) & 64.44 (116) & 20.28 (73)\\
 & Constraint & 85.83 (309) & 78.89 (142) & 40.28 (145)\\
 & Overall & 86.85 (938) & 77.41 (418) & 31.57 (341)\\
\midrule
Explicit JSON & Preference & 97.22 (350) & 94.44 (170) & 96.11 (346)\\
 & Objective & 84.44 (304) & 70.00 (126) & 84.44 (304)\\
 & Constraint & 98.61 (355) & 97.22 (175) & 92.50 (333)\\
 & Overall & 93.43 (1,009) & 87.22 (471) & 91.02 (983)\\
\bottomrule
\end{tabular}
\end{table}
Claude uses the identical task subset and four-call ordering, but quota
interruptions introduce gaps across recovery batches. Three quota-failed H5
attempts are retained separately; all 2,160 accepted responses have unique
IDs and sessions and pass prompt-hash, model, and completion-event checks.
The same fixed H5 scorer used for Astra is retained, with positive and
negative reference checks, score replay, and independent count verification.
In the order correct-to-correct, incorrect-to-correct, correct-to-incorrect,
and incorrect-to-incorrect, decision transitions are 909, 100, 29, and 42;
pair transitions are 394, 77, 24, and 45. FC transitions are 321, 662, 20,
and 77. One original response and three explicit-JSON responses are
unparseable and remain incorrect. The FC increase is therefore not principally
recovery from unparseable JSON. FC follows the H5 acceptance rules in
Appendix~\ref{app:scoring}.

% Begin inlined appendix: gemini_h5_results.tex
\FloatBarrier
\subsection{Gemini-3.1-Pro Replication}
\label{app:gemini-h5}
The Gemini-3.1-Pro replication uses the identical fixed H5 subset and randomized
four-call blocks: 1,080 logical tasks, 540 pairs, and two output conditions
(2,160 responses across 36 cells), with no tools. These records are not pooled
with earlier Gemini collections. The comparison concerns output requirements
within Gemini-3.1-Pro.

\begin{table}[ht]
\centering\small
\caption{Gemini-3.1-Pro H5. Entries are percentages (correct counts). Each source
has 360 tasks and 180 pairs per condition; overall denominators are 1,080
tasks and 540 pairs. FC uses the inherited strict H5 checker.}
\label{tab:gemini-h5}
\begin{tabular}{llrrr}
\toprule
Requirements & Source & Accuracy & IFC & FC\\
\midrule
Original & Preference & 99.17 (357) & 98.33 (177) & 61.39 (221)\\
 & Objective & 96.11 (346) & 92.22 (166) & 51.39 (185)\\
 & Constraint & 91.11 (328) & 82.22 (148) & 61.94 (223)\\
 & Overall & 95.46 (1,031) & 90.93 (491) & 58.24 (629)\\
\midrule
Explicit JSON & Preference & 100.00 (360) & 100.00 (180) & 100.00 (360)\\
 & Objective & 95.83 (345) & 91.67 (165) & 95.83 (345)\\
 & Constraint & 99.17 (357) & 98.33 (177) & 91.94 (331)\\
 & Overall & 98.33 (1,062) & 96.67 (522) & 95.93 (1,036)\\
\bottomrule
\end{tabular}
\end{table}

Explicit JSON improves aggregate Accuracy by 2.87 percentage points, IFC by
5.74, and FC by 37.69. There are 38 incorrect-to-correct decisions and seven
reversals; the corresponding pair counts are also 38 and seven. For FC, 412
responses improve and five deteriorate. The gains are not universal:
objective Accuracy declines from 346/360 to 345/360. Matching Accuracy rises
from 231/270 to 264/270, whereas apartment Accuracy is unchanged at 267/270
and meetings and allocation each lose one correct decision. Thus the
aggregate pattern replicates without a uniform source- or structure-level gain.

All 2,160 unique accepted answers pass prompt-hash, fixed-request, reported-model,
no-tool, nonempty-answer, and terminal-STOP checks; all are parseable. Thirty
failed requests remain archived separately, and recovery spans
multiple batches. No answer is selected by correctness. The original H5
scoring rules, including the stable-matching relaxation/action limitation,
are retained. A narrow type guard rejects invalid nested question objects in
five original-condition answers without changing their decision credit or
repairing their content. Replaying 4,320 existing Astra/Claude H5 scores gives
identical results; oracle controls and all new score records are also checked.
With all answers parseable, the FC increase concerns the required answer content.
\FloatBarrier
% End inlined appendix: gemini_h5_results.tex

\section{Output Requirements and Source-Specific Results}
\label{app:output}
The H2 conditions preserve the public task facts and differ in the response
requirements. The original condition retains the fixed V3 prompt, which may
already request structured output without giving every required field detail.
Thus the original condition does not mean that no JSON was requested.

Explicit JSON requires the fields \texttt{decision}, \texttt{action},
\texttt{question}, and \texttt{relax\_to}. The decision must be one of
\texttt{ACT}, \texttt{ASK\_PREFERENCE}, \texttt{ASK\_OBJECTIVE}, or
\texttt{RELAX}. A preference question identifies the agent and the two
alternatives in \texttt{compare}; an objective question lists admissible
objectives in \texttt{choose\_between}. Action and relaxation representations
are specified by structure. A relaxation must return a valid post-change action
at the top level, not only inside the correction description.

The evidence condition adds a \texttt{certificate} field. Preference evidence
provides actions under both completions; objective evidence provides an accepted
action for every admissible objective. For an \act decision these witnesses
must agree with the top-level robust action. For objective clarification,
different witnesses alone do not prove that the complete optimal sets have
empty intersection; the gold decision is checked independently. Constraint
evidence describes a valid action or the original infeasibility using the
domain-specific representation. All evidence is generated by the model, not
copied from a private oracle into its prompt.

\begin{table}[h]
\centering\small
\caption{Output-requirement comparison with counts, corresponding to Table~\ref{tab:h2}. For each reported model, each condition
contains 2,160 tasks and 1,080 pairs. Percentages appear above correct counts
in parentheses. Evidence is checked separately from FC.}
\label{tab:h2-counts}
\setlength{\tabcolsep}{3pt}
\begin{tabular}{lrrrrrr}
\toprule
& \multicolumn{3}{c}{GPT-6-Astra}
& \multicolumn{3}{c}{Claude-5-Sonnet}\\
\cmidrule(lr){2-4}\cmidrule(lr){5-7}
Requirements & Accuracy & IFC & FC & Accuracy & IFC & FC\\
\midrule
Original & \shortstack[r]{92.59\\(2,000)} & \shortstack[r]{85.74\\(926)} & \shortstack[r]{50.79\\(1,097)} & \shortstack[r]{85.74\\(1,852)} & \shortstack[r]{74.07\\(800)} & \shortstack[r]{32.18\\(695)}\\
\addlinespace
Explicit JSON & \shortstack[r]{94.58\\(2,043)} & \shortstack[r]{89.17\\(963)} & \shortstack[r]{94.40\\(2,039)} & \shortstack[r]{90.56\\(1,956)} & \shortstack[r]{81.67\\(882)} & \shortstack[r]{87.13\\(1,882)}\\
\addlinespace
JSON + evidence & \shortstack[r]{97.59\\(2,108)} & \shortstack[r]{95.19\\(1,028)} & \shortstack[r]{97.45\\(2,105)} & \shortstack[r]{94.21\\(2,035)} & \shortstack[r]{88.70\\(958)} & \shortstack[r]{91.81\\(1,983)}\\
\bottomrule
\end{tabular}
\end{table}

\begin{table}[h]
\centering\small
\caption{GPT-6-Astra H2 results by source. Each row contains 720 tasks and 360 pairs.
Percentages are followed by correct counts.}
\begin{tabular}{llrrr}
\toprule
Requirements & Source & Accuracy & IFC & FC\\
\midrule
Original & Preference & 95.28 (686) & 92.22 (332) & 58.61 (422)\\
 & Objective & 91.11 (656) & 82.22 (296) & 52.50 (378)\\
 & Constraint & 91.39 (658) & 82.78 (298) & 41.25 (297)\\
\midrule
Explicit JSON & Preference & 99.72 (718) & 99.44 (358) & 99.31 (715)\\
 & Objective & 94.31 (679) & 88.61 (319) & 94.31 (679)\\
 & Constraint & 89.72 (646) & 79.44 (286) & 89.58 (645)\\
\midrule
JSON + evidence & Preference & 99.44 (716) & 98.89 (356) & 99.31 (715)\\
 & Objective & 100.00 (720) & 100.00 (360) & 100.00 (720)\\
 & Constraint & 93.33 (672) & 86.67 (312) & 93.06 (670)\\
\bottomrule
\end{tabular}
\end{table}

The core scoring rule is identical across H2 conditions. Evidence validity
does not enter FC: otherwise the third condition would be judged
against a stricter target. For Astra, all responses in the three conditions
are parseable, the two explicit
conditions satisfy the checked field structure on every task, and 2,094 of
2,160 evidence responses pass their separate evidence checks.

% Begin inlined appendix: gemini_h2_results.tex
\subsection{Gemini-3.1-Pro Same-Task Output Requirements}
\label{app:gemini-h2}
The completed replication contains 6,480 unique answers to the same fixed
2,160 logical tasks under three requirements. Each condition covers 36 cells,
360 complete six-task groups, and 1,080 matched pairs. The original task text
and system instruction were checked against the fixed collection. Every
recorded request contains only the system instruction and the current task;
accepted responses are nonempty, end normally, and contain no tool calls.

\begin{table}[h]
\centering\small
\caption{Gemini-3.1-Pro H2 results (\%), with correct counts in parentheses.
Accuracy and FC use 2,160 tasks per condition; IFC uses 1,080 pairs.
Evidence validity is evaluated separately and does not enter FC.}
\label{tab:gemini-h2-counts}
\begin{tabular}{lrrr}
\toprule
Requirements & Accuracy & IFC & FC\\
\midrule
Original & 92.08 (1,989) & 84.35 (911) & 50.83 (1,098)\\
Explicit JSON & 95.79 (2,069) & 91.57 (989) & 95.79 (2,069)\\
JSON + evidence & 96.99 (2,095) & 93.98 (1,015) & 96.99 (2,095)\\
\bottomrule
\end{tabular}
\end{table}

FC improves by 44.95 percentage points with explicit JSON and 46.16 points
with JSON plus evidence relative to the original condition. Correct decisions
with invalid required content fall from 891 to zero in both explicit conditions.
Parsing succeeds for 2,159, 2,160, and 2,160 answers, respectively; the original
FC gap is therefore not mainly a JSON-parsing failure. Evidence is valid in
2,080 of 2,160 evidence-condition responses (96.30\%).

The verification covers all 19,457 recorded attempts: 6,480 complete answers and
12,977 failed requests, with no duplicate complete answers. Failed
attempts remain archived and are not extra scored samples. The fixed H2
scorer and oracle controls are retained. Ten malformed stable-matching
clarification fields containing unhashable nested objects are rejected as
invalid interventions without discarding the answer or removing correct
decision credit. Reported total token usage differs from the sum
of input and output tokens by one token in 504 responses; original usage
fields are retained, and failed-request costs cannot be inferred. Answer scores are computed from the recorded responses. Collection spans two
account aliases and differs in execution setup from the local Gemini
experiments.
% End inlined appendix: gemini_h2_results.tex

\section{Response-Checker Scope}
\label{app:scoring}
FC evaluates compliance with both task requirements and the specified response
interface. The V3/H5 stable-matching RELAX checker accepts the stored
repair--action pair; other valid solutions may be rejected.

For both Astra and Claude V3, verification includes positive oracle examples
for all 4,320 tasks and replay of parsed answers. Claude scoring additionally
reproduces all 4,320 archived Astra evaluations with the unchanged checkers.
H5's same-run result checks all 2,160
responses with positive examples, wrong-decision and missing-content negative
examples, and score replay. H2 checks all 6,480 responses with positive,
wrong-decision, and missing-content examples, plus 2,160 evidence positive
and negative checks and an independent count. These tests establish the
recorded computation under these acceptance rules.

\subsection{Metric Definitions}
\label{app:metric-definitions}
We separately measure decision correctness, matched-pair success, and
response-content validity, using explicit checks following verifiable
instruction-following evaluation~\citep{zhou2023ifeval}.
For $N$ responses, let $d_i=\ind[\hat y_i=y_i^\star]$ indicate a correct
decision label and $v_i$ indicate that the required content passes the fixed
checker. Let $\mathcal P_s$ be the source-$s$ pair set, with action member
$a_j$ and intervention member $b_j$. We compute
\begin{equation}
\begin{aligned}
\mathrm{Acc}&=\frac{1}{N}\sum_{i=1}^N d_i,
&\qquad \mathrm{FC}&=\frac{1}{N}\sum_{i=1}^N d_iv_i,\\
\mathrm{IFC}_s&=\frac{1}{|\mathcal P_s|}
\sum_{j\in\mathcal P_s}d_{a_j}d_{b_j},
&\qquad \mathrm{IFC}&=
\frac{\sum_s|\mathcal P_s|\mathrm{IFC}_s}{\sum_s|\mathcal P_s|}.
\end{aligned}
\end{equation}
Accuracy ($\mathrm{Acc}$) measures decision-label correctness, Fully correct
($\mathrm{FC}$) additionally requires accepted response content, and
Intervention Flip Correctness (\ifc) requires both decisions in a matched
pair to be correct. All-six accuracy requires all six decision labels in a V3
group to be correct, without requiring all six responses to pass FC.
IFC measures joint decision success (Appendix~\ref{app:independence}).

FC measures acceptance under the specified response interface and fixed
checks (Appendix~\ref{app:scoring}). Evidence
validity in the third H2 condition is scored separately and is not an
additional FC requirement. We report descriptive percentages and counts. For OSS V3, unresolved truncations remain in
the full denominator as unsuccessful deliveries, not observed reasoning
errors. Complete-response results and missing-response bounds are reported
separately; complete pair and group subsets require every member to be
complete (Appendix~\ref{app:oss-v3}).

\section{Earlier Multi-Model Experiments}
\label{app:legacy}
The earlier collection contains 150 six-task groups (900 prompts) from three
structures at $n=4$, not the four-structure V3 design. It was evaluated with
\oss, GPT-5.6-Terra, and Claude-5-Sonnet under their archived request settings.
The OSS setting used four RTX 3090 GPUs, tensor parallelism four, greedy
decoding and a 2,048-token limit. These historical
results provide supplementary comparisons; they are not scores for the
expanded OSS or Claude runs.

\begin{table}[h]
\centering\small
\caption{Earlier 900-prompt results retained from the previous manuscript.
All values are percentages. No-final answers remain failures in this historical
evaluation. All-six concerns decisions, not executable content.}
\begin{tabular}{lrrrrr}
\toprule
Model setting & Accuracy & IFC & All-six & FC & No-final\\
\midrule
GPT-OSS-120B & 75.3 & 58.9 & 23.3 & 16.8 & 8.7\\
GPT-5.6-Terra & 87.4 & 75.1 & 40.7 & 22.8 & 0.0\\
Claude-5-Sonnet & 81.1 & 69.3 & 39.3 & 2.2 & 8.6\\
\bottomrule
\end{tabular}
\end{table}
The source--structure independence references are 59.4\%, 75.2\%, and 68.4\%,
respectively, giving observed-minus-reference differences of $-0.5$, $-0.1$,
and $+0.9$ percentage points. Their small magnitude reinforces the distinction
between joint success and an additional pair dependence. In objective pairs,
both members receive the intervention response in 44.0\%, 45.3\%, and 20.7\%
of pairs, respectively.

\begin{table}[h]
\centering\small
\caption{Earlier local-cue experiment: 300 prompts and 150 pairs per model,
50 pairs per source. Percentages; distinct from the expanded H5 experiments.}
\begin{tabular}{lrrrr}
\toprule
Model setting & Overall IFC & Preference IFC & Objective IFC & Constraint IFC\\
\midrule
GPT-OSS-120B & 68.0 & 96.0 & 8.0 & 100.0\\
GPT-5.6-Terra & 80.0 & 88.0 & 82.0 & 70.0\\
Claude-5-Sonnet & 77.3 & 96.0 & 36.0 & 100.0\\
\bottomrule
\end{tabular}
\end{table}
The source ordering differs across settings. Across three independently generated
900-task OSS collections (2,700 responses), Accuracy is 75.3\%, 75.6\%,
and 75.1\%; IFC is 58.9\%, 60.4\%, and 60.2\%. These are new-instance
replications within a fixed setting, not repeated decoding of the same tasks.

\section{Earlier Output and Infeasibility Comparisons}
\label{app:legacy-format}
The earlier OSS output experiment used 360 logical tasks under three
requirements (1,080 responses). Its decision trend differs from the expanded
Astra experiment.
\begin{table}[h]
\centering\small
\caption{Earlier OSS output comparison. Percentages on 360 tasks and 180
pairs per condition; not the expanded 6,480-response Astra comparison.}
\begin{tabular}{lrrr}
\toprule
Requirements & Accuracy & IFC & FC\\
\midrule
Original & 77.2 & 62.2 & 16.1\\
Explicit JSON & 74.7 & 63.3 & 71.4\\
JSON + evidence & 71.1 & 56.1 & 62.2\\
\bottomrule
\end{tabular}
\end{table}
JSON improves execution in that setting without an aggregate decision gain.
Its JSON answers comprise 257 exact successes, four predefined
structural near misses, 38 decision errors, eight correct decisions with
invalid content, and 53 response-production failures. Near misses account for
four of 103 FC failures. Increasing the original completion cap
from 2,048 to 3,072 tokens on the same 360 tasks reduces no-final responses by
7.2 points and length finishes by 8.6 points; Accuracy rises by 7.8 points,
IFC by 9.4, and FC by 1.9. Generation budget affects both response delivery and correctness.

An earlier OSS feasibility stress test crossed two seeds with $n=4,6,8$,
using 20 feasible--infeasible pairs per seed--size combination (240 tasks).
False relaxation on the feasible members is 7.5\%, 62.5\%, and 90.0\% by size;
the $n=8$ minus $n=4$ change is +80 and +85 points within the two seeds.
These results concern the matching-with-ties task family. In the preceding
300-task exploratory collection, size and seed varied together.

\subsection{Claude First-Response Output Budget}
\label{app:claude-first-budget}
We evaluate Claude-5-Sonnet on the same 360 historical logical tasks under
first-response output caps of 2,048 and 3,072 tokens (720 independent
sessions). This three-domain collection is separate from V3. The original
task and system instructions are fixed, and cap order alternates across
tasks. Only the first complete raw message is scored with the historical
checker. Both normal and \texttt{max\_tokens} termination are retained,
including empty or malformed visible text; subsequent automatic continuations
are never merged into the scored answer. Output tokens include thinking,
so these are first-response caps, not total per-task budgets.

\begin{table}[htbp]
\centering\small
\caption{Claude first-response budget comparison. Each cap uses 360 tasks
and 180 pairs. Scores are percentages with correct counts; cap stops and
empty visible answers remain in all behavioral denominators.}
\label{tab:claude-first-budget}
\begin{tabular}{rrrrrr}
\toprule
Cap & Accuracy & IFC & FC & Cap stops & Empty text\\
\midrule
2,048 & 80.28 (289/360) & 72.22 (130/180) & 1.39 (5/360) & 55/360 & 50/360\\
3,072 & 85.83 (309/360) & 76.11 (137/180) & 1.11 (4/360) & 23/360 & 21/360\\
\bottomrule
\end{tabular}
\end{table}

Increasing the cap raises Accuracy by 5.56 percentage points: 36 incorrect
decisions become correct and 16 correct decisions become incorrect. IFC
rises by 3.89 points, while FC changes from five to four successes (two
improvements and three reversals). Thus a larger first-response budget reduces
cap termination and empty answers without improving historical FC in this
sample. FC uses the historical response-content checker, with these scores
reported separately from V3/H2 (Appendix~\ref{app:scoring}).

\begin{table}[htbp]
\centering\small
\caption{Recorded Claude output-token consumption, including thinking.
Continuation tokens are accounted for but excluded from answer scoring.}
\label{tab:claude-first-budget-usage}
\begin{tabular}{rrrrr}
\toprule
Cap & First responses & Continuations & Total & Sessions with continuation\\
\midrule
2,048 & 333,330 & 209,143 & 542,473 & 55/360\\
3,072 & 373,045 & 112,057 & 485,102 & 23/360\\
\bottomrule
\end{tabular}
\end{table}

All 720 first responses were reconstructed from complete raw message events
and reconciled with saved answers and usage. Fixed prompts and client hashes,
model requests, independent sessions, and absence of tools were checked.
The records contain 642 normal terminations and 78 cap terminations, with
continuation usage accounted for separately. Oracle positive and wrong-decision
negative controls on all 360 tasks and replay of archived historical scores
passed. No completed answer was replaced based on correctness.

\section{Other Historical Analyses}
\label{app:other}
These supplementary analyses examine task properties, feasibility at different
sizes, analysis order, and explicit-set intersection judgments.

\paragraph{Task properties.}
The previous analysis of ties reports 94.7\% raw Accuracy for tie-free
observations and 36.5\% for tie-rich observations; equal weighting of its 18
model--source--structure groups gives a difference of $-51.6$ points.
This comparison describes an association within the sampled tasks.

\paragraph{Additional checks.}
A 480-task Terra test of universal intersection versus majority or pairwise
overlap yields correct decisions throughout its eight designed groups. This
negative result bounds a particular explanation within that sample.
The Claude replication below (Appendix~\ref{app:claude-h7}) additionally
reveals unnecessary preference clarification on nonempty-intersection tasks.
Earlier analysis-first additions change OSS Accuracy only modestly:
+0.8 points relative to explicit JSON, and +3.06 points relative to an
output-matched decision-first variant; the corresponding FC changes
are $-48.9$ and $-1.94$ points. The added analysis yields small Accuracy gains and lower FC.

% Begin inlined appendix: claude_history_results.tex
\subsection{Claude Replications of Historical Comparisons}
\label{app:claude-history}
Two additional comparisons contribute 1,320 fresh Claude-5-Sonnet responses:
240 feasibility/scale tasks and 1,080 analysis-order
responses, using independent sessions without tools, MCP, or skills.
These fixed historical datasets differ
from V3 and from the earlier archived Claude API batch; their denominators
are kept separate. The separately scored first-response budget comparison
is reported in Appendix~\ref{app:claude-first-budget}.

\paragraph{Feasibility and size.}
The same matching-with-ties stress panel crosses two fixed generator seeds
with $n=4,6,8$, with 20 feasible--infeasible pairs per seed--size cell.
Table~\ref{tab:claude-history-scale} distinguishes false relaxation on feasible
tasks from RELAX recall on infeasible tasks. At $n=6,8$, the model selects
RELAX on nearly every feasible task as well as every infeasible task;
approximately 50\% Accuracy consequently does not indicate successful
feasibility discrimination. Twelve unparseable responses remain failures.
The size comparison covers these two seeds in the matching-with-ties family. Joint pair accuracy here is the
fraction of feasible--infeasible pairs with both decisions correct, not V3 IFC.

\begin{table}[ht]
\centering\small
\caption{Claude feasibility/scale panel. Percentages; Accuracy and FC use 40
tasks per row. False RELAX, RELAX recall, and joint pair accuracy each use
20 feasible tasks, infeasible tasks, or pairs, respectively.}
\label{tab:claude-history-scale}
\begin{tabular}{lrrrrrr}
\toprule
Seed & $n$ & Accuracy & FC & False RELAX & RELAX recall & Joint pair\\
\midrule
20260821 & 4 & 80.00 & 35.00 & 5.00 & 65.00 & 65.00\\
 & 6 & 50.00 & 0.00 & 100.00 & 100.00 & 0.00\\
 & 8 & 50.00 & 0.00 & 100.00 & 100.00 & 0.00\\
20260822 & 4 & 82.50 & 25.00 & 10.00 & 75.00 & 70.00\\
 & 6 & 52.50 & 0.00 & 95.00 & 100.00 & 5.00\\
 & 8 & 50.00 & 0.00 & 100.00 & 100.00 & 0.00\\
\bottomrule
\end{tabular}
\end{table}

\paragraph{Analysis before or after the decision.}
The same 360 historical tasks are evaluated under analysis-first,
output-matched decision-first, and a canonical-JSON auxiliary-field control
(1,080 responses). The original FC definition requires the correct decision,
executable intervention, and the required schema and analysis/auxiliary
fields.
All responses are parseable. Analysis-first does not improve any aggregate
metric over decision-first (Table~\ref{tab:claude-history-diagnosis}): 19
decisions improve and six deteriorate when switching to decision-first;
the FC transition counts are 23 improvements and eight reversals. The conditions vary the requested output order.

\begin{table}[ht]
\centering\small
\caption{Claude analysis-order comparison. Percentages (correct counts),
with 360 tasks and 180 pairs per condition; FC includes required extra fields.}
\label{tab:claude-history-diagnosis}
\begin{tabular}{lrrr}
\toprule
Condition & Accuracy & IFC & FC\\
\midrule
Analysis-first & 86.94 (313) & 75.56 (136) & 21.94 (79)\\
Decision-first & 90.56 (326) & 81.67 (147) & 26.11 (94)\\
Auxiliary-field control & 90.28 (325) & 80.56 (145) & 26.39 (95)\\
\bottomrule
\end{tabular}
\end{table}

\paragraph{Validation.}
The 1,320 records reported here belong to a verified collection of 4,020 unique
records and sessions, including original prompt hashes and complete response
records. The historical verifier is retained; 1,320 archived comparison rows reproduce their decision,
intervention, and FC scores, with oracle positive/negative controls on 2,220
source entries. Parsing and schema failures remain in the denominators.
Values are task-level percentages under the counting conventions defined above.
% End inlined appendix: claude_history_results.tex

\subsection{Global Intersection versus Majority and Pairwise Overlap}
\label{app:claude-h7}
We evaluate Claude-5-Sonnet on the same fixed 480-task
collection as the historical Terra comparison. Each task explicitly supplies
the complete admissible-action set for every possible world. The required
decision is \act when the intersection over all worlds is nonempty, and
\askobj otherwise. There are eight conditions, each containing 30 matched
pairs: one task with a common action and one with an empty intersection.
In the first seven conditions, $k$ denotes the number of worlds and $d$
the number that exclude the otherwise shared action. The final condition
contrasts pairwise overlap without a global intersection against a task with
a nonempty global intersection. This abstract set comparison removes the
need to derive admissible actions from a natural-language decision problem;
it is not the four-structure V3 collection.

\begin{table}[htbp]
\centering\small
\caption{Claude-5-Sonnet on the global-intersection comparison. Entries are
correct decisions out of 30 tasks per column and condition. FC equals
Accuracy in every condition. The nonempty-intersection errors are all
unnecessary \askpref decisions, not malformed JSON.}
\label{tab:claude-h7}
\begin{tabular}{lrr}
\toprule
Condition & Nonempty intersection (\act) & Empty intersection (\askobj)\\
\midrule
$k=2,\ d=1$ & 30/30 & 30/30\\
$k=3,\ d=1$ & 30/30 & 30/30\\
$k=5,\ d=1$ & 30/30 & 30/30\\
$k=8,\ d=1$ & 30/30 & 30/30\\
$k=5,\ d=2$ & 30/30 & 30/30\\
$k=8,\ d=2$ & 30/30 & 30/30\\
$k=8,\ d=4$ & 30/30 & 30/30\\
Pairwise-overlap comparison & 1/30 & 30/30\\
\midrule
Total & 211/240 & 240/240\\
\bottomrule
\end{tabular}
\end{table}

Overall Accuracy and FC are both 93.96\% (451/480), and both decisions are
correct in 211/240 pairs. All 240 empty-intersection tasks are answered
correctly: this sample provides no evidence of false \act decisions caused
by replacing global intersection with majority agreement or pairwise overlap.
Instead, all 29 errors occur in the nonempty-intersection counterparts of
the pairwise-overlap condition. These tasks already admit multiple common
actions, yet Claude selects \askpref rather than any permitted action.
The errors consist of unnecessary clarification when several valid actions remain.

\paragraph{Record verification and client retries.}
All 480 original prompt hashes, unique sessions, returned model names,
no-tool settings, and complete final responses were verified before scoring
with the unchanged checker. The client
logged 49 internal retry events across 37 tasks, all before the first observed
assistant output. Every task has exactly one final text response, with no
runner-level replacement calls or correctness-based selection. Keeping all
480 tasks is the primary analysis. On the 443 tasks without a recorded
retry, Accuracy and FC are both 93.91\% (416/443); this subset is not a
balanced matched design. Raw retry events are retained, and the logs cannot
exclude unobserved server-side computation before a retry.

\section{Task Completion after Intervention}
\label{app:completion}
\paragraph{Selection and scale.}
We sample five of the ten matched groups available in each cell of the fixed
H2 source collection, using seed 20260915 and without consulting model
outcomes. Only the three intervention-needed members of each group are used:
$36\times5\times3=540$ tasks, with 180 per source, 135 per structure, and
180 per size. The 36 cells cover four structures, $n\in\{4,6,8\}$, and
generator seeds 20260901, 20260902, and 20260903. A 12-task initial batch
(one per structure and source, at $n=4$ and the first generator seed) was
checked before the remaining 528 tasks were generated. Its results are
retained without replacement for Astra. Claude uses the same 540 tasks and
feedback rules; its two explicitly authorized fresh attempts are documented below.
Gemini also uses these 540 tasks and the same fixed feedback and scoring rules.

\paragraph{Routes and feedback.}
All three models run both routes. In the
autonomous route, the first response must contain the four JSON fields
\texttt{decision}, \texttt{action}, \texttt{question}, and \texttt{relax\_to},
with \texttt{action} set to null until feedback is received. A preference
question must identify the missing comparison; an objective question must
list all admissible objectives; a relaxation must propose a permitted minimum
change. A valid request receives feedback, followed by a final-action request
in the same conversation. An invalid first response is an experimental
failure: no corrective hint, second turn, or retry is supplied.

Preference completions and selected objectives are fixed by the sampling seed.
For relaxation, feedback approves the proposed minimum change and retains all
other requirements. The final action is checked against the source certificate
for that updated task. The reference route uses a separate conversation and
directly supplies the predetermined feedback; for relaxation, it uses the first
minimum correction in a fixed ordering. Route order alternates across tasks,
while each autonomous dialogue remains contiguous. The autonomous and
reference routes can have different valid relaxation choices and
different conversation histories.

\paragraph{Completion measures.}
First-step validity measures whether the required intervention is correctly
specified. Autonomous end-to-end success requires both a valid intervention
and a correct final action after feedback. Reference success requires a correct
final action given the reference feedback. All tasks remain in their respective
denominators, including those without a valid first intervention. Final answers
must use the four specified fields, choose \texttt{ACT}, provide a certified
valid optimal action (or a stable matching when required), and set the question
and relaxation fields to null.

\begin{table}[htbp]
\centering\small
\caption{Astra completion by source. Entries are correct tasks over all tasks in the
row. The overall percentages are 99.44\%, 99.07\%, and 99.81\%, respectively.}
\label{tab:completion-source}
\begin{tabular}{lrrr}
\toprule
Source & Valid intervention & Autonomous completion & Reference completion\\
\midrule
Preference & 178/180 & 177/180 & 180/180\\
Objective & 179/180 & 179/180 & 179/180\\
Constraint & 180/180 & 179/180 & 180/180\\
\midrule
Overall & 537/540 & 535/540 & 539/540\\
\bottomrule
\end{tabular}
\end{table}

\begin{table}[htbp]
\centering\small
\caption{Astra completion by structure and size. These are alternative breakdowns
of the same 540 tasks, not additional experiments.}
\label{tab:completion-strata}
\begin{tabular}{lrr}
\toprule
Subset & Autonomous completion & Reference completion\\
\midrule
Object allocation & 133/135 & 134/135\\
Meeting scheduling & 135/135 & 135/135\\
Apartment choice & 133/135 & 135/135\\
Stable matching & 134/135 & 135/135\\
\midrule
$n=4$ & 179/180 & 180/180\\
$n=6$ & 179/180 & 180/180\\
$n=8$ & 177/180 & 179/180\\
\bottomrule
\end{tabular}
\end{table}

\paragraph{Astra results and all failures.}
Conditional on a valid first intervention, autonomous completion is 535/537
(99.63\%); the unconditional rate is 535/540 (99.07\%). Both routes succeed
on 534 tasks, only the reference succeeds on five, and only the autonomous
route succeeds on one. The reference advantage of four tasks (0.74 percentage
points) is descriptive. All six failed route responses are parseable four-field
JSON; they are not syntax failures.

The three invalid interventions are \texttt{CMPB0099} (stable matching,
objective, $n=8$), \texttt{CMPB0245} (apartment choice, preference, $n=8$), and
\texttt{CMPB0398} (allocation, preference, $n=6$). Each chooses \texttt{ACT}
instead of requesting the missing information. Two further tasks have valid
interventions but wrong final actions: \texttt{CMPB0001} (apartment choice,
preference, $n=4$) selects Apartment2 after feedback, whereas the reference
selects the correct Apartment4; \texttt{CMPB0306} (allocation, constraint,
$n=8$) proposes a valid minimum change but returns an allocation outside the
corresponding accepted set. Its reference route uses a different permitted
change, so the comparison is not on an identical updated task. Finally,
\texttt{CMPB0483} (allocation, objective, $n=8$) succeeds autonomously after
objective clarification but gives a wrong allocation in the reference route.

\paragraph{Astra record checks.}
The expected maximum is 1,620 calls. The three invalid interventions correctly
omit their second turn, leaving 1,617 completed calls across 1,080 route
conversations. All records pass checks for unique task and turn IDs, fixed
input hashes, expected prompts and feedback, conversation continuity and
separation, completion events, final-response agreement, and absence of tool
items. Recomputed outcomes match all saved task results. Feedback sets are checked against the fixed source certificates.

The collection uses previously selected synthetic intervention-needed tasks,
with autonomous and direct-feedback reference routes.
The fixed runner checks model identity and additional instruction sources at
conversation creation, but those Astra creation receipts were not saved. Provider-side model versions are identified by the requested model names.
\paragraph{Claude completion results.}
Claude has 427 valid first interventions and 408 autonomous completions,
versus 516 reference completions, out of 540 tasks
(79.07\%, 75.56\%, and 95.56\%). Conditional on a valid first step,
completion is 408/427 (95.55\%). Both routes succeed on 395 tasks;
only the reference succeeds on 121 and only the autonomous route on 13.
The 20.00-percentage-point aggregate difference is descriptive: the routes
differ in history and can use different permitted constraint changes.

\begin{table}[htbp]
\centering\small
\caption{Claude completion by source, using the same 540 tasks and fixed
feedback rules as Astra. Entries are successful tasks over all tasks.}
\label{tab:claude-completion-source}
\begin{tabular}{lrrr}
\toprule
Source & Valid intervention & Autonomous completion & Reference completion\\
\midrule
Preference & 171/180 & 166/180 & 170/180\\
Objective & 163/180 & 152/180 & 171/180\\
Constraint & 93/180 & 90/180 & 175/180\\
\midrule
Overall & 427/540 & 408/540 & 516/540\\
\bottomrule
\end{tabular}
\end{table}

Of 113 invalid first responses, 74 provide an action before feedback
(all constraint tasks), 22 fail the strict JSON shape, nine choose the wrong
decision, and eight omit part of the objective question. These categories
follow the checker's fixed order and are mutually exclusive, not a count of
every possible defect. Nineteen tasks fail the final-action check after a
valid intervention. Thus the low autonomous score on constraints partly
reflects failure to follow the required two-step procedure, not solely
inability to produce an action after an approved repair.

\begin{table}[htbp]
\centering\small
\caption{Claude completion by structure and size: alternative breakdowns
of the same 540 tasks.}
\begin{tabular}{lrr}
\toprule
Subset & Autonomous completion & Reference completion\\
\midrule
Object allocation & 92/135 & 116/135\\
Meeting scheduling & 114/135 & 135/135\\
Apartment choice & 125/135 & 134/135\\
Stable matching & 77/135 & 131/135\\
\midrule
$n=4$ & 138/180 & 178/180\\
$n=6$ & 134/180 & 178/180\\
$n=8$ & 136/180 & 160/180\\
\bottomrule
\end{tabular}
\end{table}

\paragraph{Claude recovery and checks.}
There are 1,507 accepted response turns in 1,080 independent route sessions:
540 first steps, 427 autonomous follow-ups, and 540 reference answers.
Invalid first steps deliberately receive no follow-up and remain failures
in the 540-task denominator. Exact feedback, real conversation continuity,
model requests, no-tool settings, and terminal events were checked.
The unchanged completion scoring rules reproduce all 540 stored Astra outcomes.
Bare JSON is required; no code fences or answers are repaired before scoring.

Two interrupted tasks, \texttt{CMPB0001} and \texttt{CMPB0197}, had complete
first responses but no successful follow-up and no resumable original session.
The owner explicitly authorized fresh full-task attempts, including repeated
first steps. These are marked as restarts, not continuations; original partial
records are preserved and selection does not depend on correctness.
Excluding these two tasks gives 425/538 valid interventions (79.00\%),
406/538 autonomous completions (75.46\%), and 514/538 reference completions
(95.54\%). One archived reference answer for \texttt{CMPB0228} was reused
after validation of a client retry event before answer output; it was not
called again. The logs cover client-side requests and recovery events.

\paragraph{Gemini completion results.}
Under the unchanged bare-JSON rules, Gemini has 39/540 valid first
interventions (7.22\%), 6/540 autonomous completions (1.11\%), and
29/540 reference completions (5.37\%; Table~\ref{tab:gemini-completion-source}).
Among the 39 tasks with a valid first step, six pass the final check
(15.38\%). Only the reference route succeeds on 29 tasks, only the autonomous
route on six, and both on none; the aggregate difference is 4.26 percentage
points, calculated from the unrounded counts.

\begin{table}[htbp]
\centering\small
\caption{Gemini completion by source under the fixed bare-JSON requirements.
All tasks remain in the denominators, including those rejected before feedback.}
\label{tab:gemini-completion-source}
\begin{tabular}{lrrr}
\toprule
Source & Valid intervention & Autonomous completion & Reference completion\\
\midrule
Preference & 10/180 & 1/180 & 5/180\\
Objective & 12/180 & 4/180 & 10/180\\
Constraint & 17/180 & 1/180 & 14/180\\
\midrule
Overall & 39/540 & 6/540 & 29/540\\
\bottomrule
\end{tabular}
\end{table}

All 501 invalid first responses contain four-field JSON inside Markdown
code fences. The fixed parser applies JSON decoding to the entire response
without stripping fences, so these responses do not trigger feedback or a
second turn. Code fences also account for all 33 rejected autonomous final
responses and all 511 rejected reference responses. Content scoring applies to responses accepted by the bare-JSON parser.
The 6/39 conditional rate covers the parser-accepted first responses;
the other 501 tasks end before feedback.

\paragraph{Gemini collection and checks.}
The collection contains 1,119 complete response turns in 1,080 independent
route sessions: 540 first steps, 39 autonomous follow-ups, and 540 reference
answers. Resumed collection preserves complete stages and their actual
conversation history; six unsuccessful attempts remain archived and are
excluded from response counts. Prompt and fixed-configuration hashes,
feedback branches, unique terminal events, nonempty responses, reported
model identity, and absence of tool calls were checked for every accepted
stage, with no duplicate complete response or unaggregated answer found.
The scoring checks pass 4,227 positive and negative controls and reproduce
all 540 stored Astra and all 540 stored Claude task outcomes.

\section{Answer Selection and Intervention Wording}
\label{app:selection-framing}
\paragraph{Shared sample and status.}
Both experiments are complete for GPT-6-Astra, Claude-5-Sonnet, and
Gemini-3.1-Pro. Each model contributes 1,080 selection responses and 2,160
wording responses (1,080 per condition), totaling 3,240 new responses verified
and scored. The generation baseline reuses each model's own same-task H2
explicit-JSON responses.
We sample five of the ten complete six-task groups
per cell in the H2 collection using a fixed sampling seed (20260916), without
reading model answers or scores. The sample contains $36\times5\times6=1,080$
tasks, 540 action/intervention pairs, and 180 complete groups. Each source
contributes 360 tasks and 180 pairs. Astra uses
the same setup as its completed H2 comparison.
For Astra and Claude, each task receives one selection call and two wording calls, totaling 3,240
new calls. The three calls are ordered randomly within each task block and
use separate fresh conversations. A 12-call initial batch is checked for
record integrity before the remaining 3,228 calls; answer correctness is not
a criterion for expanding the run or retaining tasks.

\subsection{Generating versus Selecting an Answer}
For each task, we reuse its existing H2 explicit-JSON generation response as
the baseline and request one new four-option selection response. Each menu
contains two ACT responses and two interventions of the relevant source,
with exactly one fully correct candidate under the reference checker. All
four candidates satisfy the required JSON field types. Correct positions
are balanced globally: 270 each at A, B, C, and D. The model returns a JSON
object with a single \texttt{choice} field containing A, B, C, or D; scoring
maps the selected candidate to its decision and content. An invalid selection
remains an incorrect answer, not a reason to
repeat the call. Accuracy, IFC, and FC retain their logical definitions.

The preference distractor asks about a known comparison; the objective
distractor repeats an admissible objective instead of distinguishing the
alternatives; the relaxation distractor retains a minimum change but supplies
an invalid resulting action. Supplying candidates reduces the construction
burden and provides answer content, while distractors vary in difficulty.
We report paired improvements and reversals,
not only aggregate differences. Generation-format compliance and selection-output
compliance are not treated as equally difficult measures.

\begin{table}[htbp]
\centering\small
\caption{Generation versus selection: GPT-6-Astra.
Each condition has 1,080 tasks and 540 pairs overall, or 360 tasks and 180
pairs per source. Percentages appear above correct counts. The generation
baseline is rescored on the same selected subset, not copied from the full H2 score.}
\label{tab:selection-pending}
\begin{tabular}{llrrr}
\toprule
Source & Response mode & Accuracy & IFC & FC\\
\midrule
All & Generate (explicit JSON) & \shortstack{94.72\\(1023/1080)} & \shortstack{89.44\\(483/540)} & \shortstack{94.63\\(1022/1080)}\\
& Select (four options) & \shortstack{97.59\\(1054/1080)} & \shortstack{95.19\\(514/540)} & \shortstack{97.59\\(1054/1080)}\\
\midrule
Preference & Generate & \shortstack{99.72\\(359/360)} & \shortstack{99.44\\(179/180)} & \shortstack{99.44\\(358/360)}\\
& Select & \shortstack{98.89\\(356/360)} & \shortstack{97.78\\(176/180)} & \shortstack{98.89\\(356/360)}\\
Objective & Generate & \shortstack{93.61\\(337/360)} & \shortstack{87.22\\(157/180)} & \shortstack{93.61\\(337/360)}\\
& Select & \shortstack{93.89\\(338/360)} & \shortstack{87.78\\(158/180)} & \shortstack{93.89\\(338/360)}\\
Constraint & Generate & \shortstack{90.83\\(327/360)} & \shortstack{81.67\\(147/180)} & \shortstack{90.83\\(327/360)}\\
& Select & \shortstack{100.00\\(360/360)} & \shortstack{100.00\\(180/180)} & \shortstack{100.00\\(360/360)}\\
\bottomrule
\end{tabular}
\end{table}

% Begin inlined appendix: gemini_selection_results.tex
\paragraph{Gemini selection replication.}
All 1,080 selection responses pass the fixed request and complete-response
checks, including four reused responses and 1,076 subsequently collected with
four concurrent workers. Three failed requests are retained separately;
there are no duplicate complete answers or unaggregated complete responses.
All outputs are valid choices. The same-task H2 baseline is linked by instance
ID and its core scores are replayed before comparison.

\begin{table}[htbp]
\centering\small
\caption{Gemini-3.1-Pro generation versus selection by source. Each source has
360 tasks and 180 pairs; Overall has 1,080 tasks and 540 pairs. Percentages
appear above correct counts. Generation reuses the same-task H2 explicit-JSON subset.}
\label{tab:gemini-selection-source}
\begin{tabular}{llrrr}
\toprule
Source & Response mode & Accuracy & IFC & FC\\
\midrule
Preference & Generate & \shortstack{98.89\\(356/360)} & \shortstack{97.78\\(176/180)} & \shortstack{98.89\\(356/360)}\\
& Select & \shortstack{98.89\\(356/360)} & \shortstack{97.78\\(176/180)} & \shortstack{98.89\\(356/360)}\\
Objective & Generate & \shortstack{90.56\\(326/360)} & \shortstack{81.11\\(146/180)} & \shortstack{90.56\\(326/360)}\\
& Select & \shortstack{91.39\\(329/360)} & \shortstack{82.78\\(149/180)} & \shortstack{91.39\\(329/360)}\\
Constraint & Generate & \shortstack{97.22\\(350/360)} & \shortstack{94.44\\(170/180)} & \shortstack{97.22\\(350/360)}\\
& Select & \shortstack{100.00\\(360/360)} & \shortstack{100.00\\(180/180)} & \shortstack{100.00\\(360/360)}\\
\midrule
Overall & Generate & \shortstack{95.56\\(1032/1080)} & \shortstack{91.11\\(492/540)} & \shortstack{95.56\\(1032/1080)}\\
& Select & \shortstack{96.76\\(1045/1080)} & \shortstack{93.52\\(505/540)} & \shortstack{96.76\\(1045/1080)}\\
\bottomrule
\end{tabular}
\end{table}

For both Accuracy and FC, 25 tasks improve and 12 reverse, giving a net gain
of 13. Preference scores are unchanged in aggregate; objective and constraint
FC gain three and ten tasks, respectively. Unnecessary interventions fall
from 45/540 to 35/540 ACT tasks; missed interventions fall from 3/540 to 0/540.
% End inlined appendix: gemini_selection_results.tex

\subsection{Neutral Wording versus an Intervention Reminder}
Both conditions retain the H2 explicit-JSON prompt. The neutral suffix reads:
\begin{quote}\small
Apply the decision rule stated in the task and select the logically appropriate
next step. Use only the supplied information.
\end{quote}
The reminder suffix reads:
\begin{quote}\small
Remember that asking for missing preferences or objectives and relaxing
infeasible constraints are available next steps when necessary. Use only the
supplied information.
\end{quote}
No task fact, gold label, or scoring requirement changes. Because the original
prompt already defines intervention choices, this tests incremental emphasis,
not whether the model knows that clarification is allowed. Each condition has
540 ACT cases and 540 necessary-intervention cases. Besides Accuracy, IFC,
and FC, we report unnecessary intervention on ACT cases and missed
intervention on cases requiring ASK or RELAX. Choosing an incorrect type of
intervention is a label error but not a missed intervention; unparseable
decisions are reported separately and remain incorrect for Accuracy. All new
Astra responses were parseable. Claude has one parsing failure in each
wording condition and one in its reused generation baseline; all 1,080
selection outputs are valid choices. We also examine paired changes and results by source,
structure, and size; each cell contributes five groups.

Figure~\ref{fig:framing-main} visualizes the aggregate results; Table~\ref{tab:framing-detail} gives exact scores and counts. For Astra, the reminder changes
16 incorrect decisions to correct ones and 12 correct decisions to incorrect
ones; for FC, the corresponding counts are 18 and 16. By source, FC changes
from 355 to 355 of 360 preference tasks, 342 to 336 objective tasks, and 325
to 333 constraint tasks. The small aggregate difference therefore combines
improvement and deterioration across sources.
For Claude, the reminder changes 47 wrong decisions to correct ones and
37 correct decisions to wrong ones; FC has 63 improvements and 48 reversals.
Source FC changes from 319 to 322 (preference), 288 to 291 (objective),
and 331 to 340 (constraint), each out of 360 tasks.
For Gemini-3.1-Pro, Accuracy and FC rise from 95.28\% to 95.93\%,
with 18 improvements and 11 reversals; IFC rises from 90.56\% to 92.04\%.
Unnecessary interventions fall from 50 to 43, while missed interventions
remain at one.
\begin{table}[htbp]
\centering\small
\caption{Claude generation versus selection by source. Each source row has 360
tasks and 180 pairs; Overall has 1,080 tasks and 540 pairs. Percentages appear above correct counts; the generation
baseline is the same-task subset of Claude H2.}
\label{tab:claude-selection-source}
\begin{tabular}{llrrr}
\toprule
Source & Response mode & Accuracy & IFC & FC\\
\midrule
Preference & Generate & \shortstack{91.11\\(328/360)} & \shortstack{83.33\\(150/180)} & \shortstack{90.00\\(324/360)}\\
& Select & \shortstack{76.94\\(277/360)} & \shortstack{54.44\\(98/180)} & \shortstack{76.94\\(277/360)}\\
Objective & Generate & \shortstack{82.22\\(296/360)} & \shortstack{65.00\\(117/180)} & \shortstack{81.11\\(292/360)}\\
& Select & \shortstack{73.61\\(265/360)} & \shortstack{47.22\\(85/180)} & \shortstack{73.61\\(265/360)}\\
Constraint & Generate & \shortstack{98.33\\(354/360)} & \shortstack{96.67\\(174/180)} & \shortstack{91.39\\(329/360)}\\
& Select & \shortstack{99.72\\(359/360)} & \shortstack{99.44\\(179/180)} & \shortstack{98.89\\(356/360)}\\
\midrule
Overall & Generate & \shortstack{90.56\\(978/1080)} & \shortstack{81.67\\(441/540)} & \shortstack{87.50\\(945/1080)}\\
& Select & \shortstack{83.43\\(901/1080)} & \shortstack{67.04\\(362/540)} & \shortstack{83.15\\(898/1080)}\\
\bottomrule
\end{tabular}
\end{table}

Unlike Astra, Claude has a net decline under selection: 29 wrong decisions
become correct but 106 correct decisions become wrong. FC has 57 improvements
and 104 reversals. Preference and objective performance declines despite
the constraint improvement. Selection increases unnecessary intervention
from 89/540 to 174/540 ACT tasks, while missed intervention falls from
12/540 to 5/540 intervention-needed tasks.

\begin{table}[htbp]
\centering\small
\caption{Neutral wording versus an intervention reminder: exact scores for Figure~\ref{fig:framing-main}. Each model uses the same 1,080 tasks and 540 pairs per condition. Percentages appear above counts. Unnecessary and missed intervention rates use 540 ACT and 540 necessary-intervention cases, respectively.}
\label{tab:framing-detail}
\setlength{\tabcolsep}{5pt}
\begin{tabular}{@{}lrrrrr@{}}
\toprule
& \multicolumn{3}{c}{Performance $\uparrow$} & \multicolumn{2}{c}{Intervention errors $\downarrow$}\\
\cmidrule(lr){2-4}\cmidrule(l){5-6}
Condition & Accuracy & IFC & FC & Unnecessary & Missed\\
\midrule
\multicolumn{6}{@{}l}{\textbf{GPT-6-Astra}}\\
\addlinespace[2pt]
Neutral & \shortstack{94.81\\(1024/1080)} & \shortstack{89.63\\(484/540)} & \shortstack{94.63\\(1022/1080)} & \shortstack{9.63\\(52/540)} & \shortstack{0.74\\(4/540)}\\
Reminder & \shortstack{95.19\\(1028/1080)} & \shortstack{90.37\\(488/540)} & \shortstack{94.81\\(1024/1080)} & \shortstack{8.89\\(48/540)} & \shortstack{0.74\\(4/540)}\\
\midrule
\multicolumn{6}{@{}l}{\textbf{Claude-5-Sonnet}}\\
\addlinespace[2pt]
Neutral & \shortstack{89.63\\(968/1080)} & \shortstack{80.19\\(433/540)} & \shortstack{86.85\\(938/1080)} & \shortstack{17.04\\(92/540)} & \shortstack{3.52\\(19/540)}\\
Reminder & \shortstack{90.56\\(978/1080)} & \shortstack{81.67\\(441/540)} & \shortstack{88.24\\(953/1080)} & \shortstack{15.93\\(86/540)} & \shortstack{2.78\\(15/540)}\\
\midrule
\multicolumn{6}{@{}l}{\textbf{Gemini-3.1-Pro}}\\
\addlinespace[2pt]
Neutral & \shortstack{95.28\\(1029/1080)} & \shortstack{90.56\\(489/540)} & \shortstack{95.28\\(1029/1080)} & \shortstack{9.26\\(50/540)} & \shortstack{0.19\\(1/540)}\\
Reminder & \shortstack{95.93\\(1036/1080)} & \shortstack{92.04\\(497/540)} & \shortstack{95.93\\(1036/1080)} & \shortstack{7.96\\(43/540)} & \shortstack{0.19\\(1/540)}\\
\bottomrule
\end{tabular}
\end{table}

\paragraph{Completion and verification.}
The final Astra record set contains 3,240 unique task IDs and fresh conversation IDs,
with exact prompt, requested model, no-tool, completed-turn,
and response/event agreement checks. Each condition covers all 36 cells,
180 complete six-task groups, and 540 action/intervention pairs. One failed
capacity-related attempt is retained but replaced by its authorized continuation.
A later local interruption left one complete response outside the running
summary; it was recovered from matching response and completion records rather
than called again. All 4,320 candidate answers were checked, including unique
correctness of each menu; standard-answer and invalid-answer tests and an
independent recount passed. No answers were selected for inclusion by correctness.
Claude likewise contains 3,240 complete new responses and reuses 1,080
same-task Claude H2 responses. Three quota-failed attempts are retained
separately, and complete answers were not called again. All records pass the
checks in Appendix~\ref{app:reproducibility}. The same scoring function reproduces all 3,240 archived
Astra supplement scores; all candidate and oracle checks pass.
The Gemini wording collection contains 2,160 unique complete responses,
with 1,080 per condition, verified under the same record checks.
The unchanged scorer reproduces the archived Astra and Claude wording scores;
all 1,080 oracle positives and both sets of negative controls pass.
Reported token totals disagree with their component sums in 201 records;
these are preserved and do not enter behavioral scoring.

\section{Inference Settings and Failure Accounting}
\label{app:reproducibility}
\paragraph{Recorded inference settings.}
Table~\ref{tab:inference-settings} reports request-side settings for the main
collections and their later supplements. ``Not set'' means that the experiment
runner did not explicitly supply the parameter; it does not identify a numerical
client or server default. Reasoning levels are requested settings, not measured
reasoning-token budgets. No explicit top-$p$, top-$k$, or frequency/presence
penalty override is supplied in these configurations.

\begin{table}[htbp]
\centering
\caption{Recorded inference configurations. Output caps are in tokens. V3, H2,
and H5 denote the main, output-requirement, and local-cue collections,
respectively. Selection, wording, and completion denote the corresponding
supplementary comparisons.}
\label{tab:inference-settings}
\begingroup
\footnotesize
\setlength{\tabcolsep}{3pt}
\renewcommand{\arraystretch}{1.13}
\begin{tabular}{@{}>{\raggedright\arraybackslash}p{0.33\linewidth}>{\raggedright\arraybackslash}p{0.31\linewidth}>{\raggedright\arraybackslash}p{0.16\linewidth}>{\raggedright\arraybackslash}p{0.13\linewidth}@{}}
\toprule
Model / collection & Access interface & Output cap & Reasoning\\
\midrule
GPT-6-Astra: V3 & OpenAI (Responses API) & Not set & medium\\
GPT-6-Astra: H2, H5, selection, wording, completion & OpenAI (Responses API) & Not set & medium\\
Claude-5-Sonnet: V3, H2, H5 and later supplements except budget comparison & Anthropic (Messages API) & Not set & medium\\
Claude-5-Sonnet: first-response budget comparison & Anthropic (Messages API) & 2,048 / 3,072 & medium\\
Gemini-3.1-Pro: H5, selection & Google (Gemini API) & Not set & high\\
Gemini-3.1-Pro: imported H2, wording & Google (Gemini API) & Not set & high\\
Gemini-3.1-Pro: V3 (AGY / API) & AGY / Gemini-compatible API & Not recorded / 65,536 & high\\
Gemini-3.1-Pro: completion & Gemini-compatible API & 65,536 & high\\
\oss: V3 & Local OpenAI-compatible server; four RTX 3090 GPUs & 2,048 / 3,072 / 6,144 & medium\\
\bottomrule
\end{tabular}
\endgroup
\end{table}

Gemini's V3 API and completion requests set temperature to 1.0.
The requested identifiers are \texttt{gpt-6-astra}, \texttt{claude-sonnet-5},
\texttt{gemini-3.1-pro}, and \texttt{openai/gpt-oss-120b}.
Claude's budget comparison sets the client output-token limit separately for
each condition and scores only the first complete message, including
\texttt{max\_tokens} terminations; it is not a cap on total session consumption
(Appendix~\ref{app:claude-first-budget}). The OSS limits are successive collection
and recovery settings rather than three independent full evaluations
(Appendix~\ref{app:oss-v3}). Earlier historical collections retain their separate
settings: the archived Claude API batches request temperature 1, a 2,048-token
cap, adaptive thinking, and medium effort; the archived Terra supplemental
requests use \texttt{gpt-5.6-terra} through OpenAI (Responses API) with medium effort.

\paragraph{Model configurations.}
Provider, inference implementation, and collection time vary across models.
Requests record model identifiers rather than fixed provider snapshots.
The primary comparisons concern observed responses within each model and
task collection. Earlier multi-model results use separate, smaller collections;
pending expansions are excluded.

The main text uses the completed, checked Astra V3 result and the later
same-run H5, full H2, and checked task-completion results,
together with the complete Claude V3, H5, and H2 replications. Both models
also have scored selection, wording, and task-completion comparisons;
the two Claude full-task restarts are explicitly separated above.
Gemini has scored replications of V3 (4,320 responses), H2 (6,480 responses),
and H5 (2,160 responses), a 1,080-task selection comparison with its same-task H2
baseline, a scored 2,160-response wording comparison, and a scored
540-task completion comparison under the strict response requirements
(Appendix~\ref{app:completion}).
Earlier separately generated
H5 original/JSON batches are not substituted for these
comparisons. They remain archived for comparison. The full OSS V3 collection
has 8,640 IDs, with 8,455 complete responses and 185 unresolved truncations.
It is scored with explicit missing-response accounting, not described as
a collection with all responses complete (Appendix~\ref{app:oss-v3}). Claude V3 contains 4,320 accepted responses
across four batches (2,721, 410, 1,109, and 80). Two quota-failed attempts
and one connection-failed attempt are preserved separately. All accepted
responses use independent sessions, record
\texttt{claude-sonnet-5}, and contain no tool calls; no accepted answer is
duplicated or selected by correctness. Claude also contributes 6,480 H2 and 2,160 H5
accepted responses; eight H2 and three H5 quota-failed attempts are archived
separately. No partial batch is substituted for either complete comparison.

Before scoring, the pipeline requires expected ID coverage, no duplicate
accepted answers, matching prompt hashes, valid completion records, and
private linkage to the intended logical tasks. Request failures are retained
in their original files, and authorized recovery is tracked separately.
For OSS, this completion requirement determines which answers enter the
complete-response subset; all intended IDs remain in full-denominator scores,
with unresolved truncation explicitly marked rather than parsed as an answer.
It does not discard completed answers because they are wrong. In the complete
Astra H2 comparison, one quota-failed attempt is retained separately while
the final evaluated collection contains one valid response for each of 6,480
IDs. Parsing failures in completed responses remain incorrect, as in the
single Astra V3 parsing failure and the single original-condition Astra H5
parsing failure. Claude H5 retains one original-condition and three
explicit-condition parsing failures in its behavioral denominators.

These behavioral denominators differ from a service-reliability denominator
that counts every failed request. The older 900-task tables also retain their
historical no-final accounting and are reported separately from the
recovered expanded runs. Archived prompts, checker
versions, request settings, timestamps, response records, and recovery mappings
are required to reproduce the reported comparisons.
\clearpage
\begin{table}[p]
\centering\small
\caption{All 36 Astra V3 cells. Each cell contains 120 tasks, 60 pairs, and
20 six-task groups. Entries are correct counts; the denominators are fixed
within each column. All-six FC is zero in every cell.}
\label{tab:v3-cells}
\begin{tabular}{lrrrrrr}
\toprule
Structure & $n$ & Seed & Accuracy/120 & IFC/60 & FC/120 & All-six/20\\
\midrule
Apartments & 4 & 20260901 & 116 & 56 & 55 & 16 \\
Apartments & 4 & 20260902 & 117 & 57 & 57 & 17 \\
Apartments & 4 & 20260903 & 118 & 58 & 59 & 18 \\
Apartments & 6 & 20260901 & 119 & 59 & 57 & 19 \\
Apartments & 6 & 20260902 & 116 & 57 & 55 & 17 \\
Apartments & 6 & 20260903 & 114 & 56 & 53 & 16 \\
Apartments & 8 & 20260901 & 118 & 58 & 56 & 18 \\
Apartments & 8 & 20260902 & 119 & 59 & 57 & 19 \\
Apartments & 8 & 20260903 & 119 & 59 & 56 & 19 \\
\midrule
Meetings & 4 & 20260901 & 115 & 56 & 56 & 16 \\
Meetings & 4 & 20260902 & 114 & 55 & 55 & 15 \\
Meetings & 4 & 20260903 & 114 & 56 & 55 & 16 \\
Meetings & 6 & 20260901 & 117 & 58 & 59 & 18 \\
Meetings & 6 & 20260902 & 119 & 59 & 57 & 19 \\
Meetings & 6 & 20260903 & 119 & 59 & 59 & 19 \\
Meetings & 8 & 20260901 & 120 & 60 & 60 & 20 \\
Meetings & 8 & 20260902 & 120 & 60 & 58 & 20 \\
Meetings & 8 & 20260903 & 118 & 58 & 59 & 18 \\
\midrule
Allocation & 4 & 20260901 & 102 & 42 & 42 & 7 \\
Allocation & 4 & 20260902 & 101 & 41 & 41 & 7 \\
Allocation & 4 & 20260903 & 104 & 44 & 44 & 7 \\
Allocation & 6 & 20260901 & 100 & 40 & 39 & 4 \\
Allocation & 6 & 20260902 & 98 & 38 & 37 & 3 \\
Allocation & 6 & 20260903 & 96 & 36 & 37 & 4 \\
Allocation & 8 & 20260901 & 93 & 33 & 34 & 2 \\
Allocation & 8 & 20260902 & 97 & 37 & 38 & 3 \\
Allocation & 8 & 20260903 & 88 & 29 & 31 & 3 \\
\midrule
Matching & 4 & 20260901 & 108 & 48 & 88 & 8 \\
Matching & 4 & 20260902 & 111 & 51 & 91 & 11 \\
Matching & 4 & 20260903 & 107 & 47 & 87 & 7 \\
Matching & 6 & 20260901 & 114 & 54 & 94 & 14 \\
Matching & 6 & 20260902 & 114 & 54 & 94 & 14 \\
Matching & 6 & 20260903 & 110 & 50 & 90 & 10 \\
Matching & 8 & 20260901 & 115 & 55 & 95 & 15 \\
Matching & 8 & 20260902 & 116 & 56 & 96 & 16 \\
Matching & 8 & 20260903 & 113 & 53 & 93 & 13 \\
\bottomrule
\end{tabular}
\end{table}
% Begin inlined appendix: protocol_inventory.tex
\clearpage
\section{Experimental Design and Validation}
\label{app:protocol-inventory}

For correspondence with the archived experiment files, V3 is indexed as H1,
the local-cue comparison as H5, and the output-requirement comparison as H2.

This overview separates logical tasks from prompt conditions and model
responses and describes the fixed designs underlying the main comparisons.
Completed results and
model coverage are reported in the corresponding result sections; historical
experiments and pending collections are not pooled into these totals.
The selection, wording, and task-completion comparisons are scored for Astra, Claude,
and Gemini.

\begin{table}[htbp]
\centering\small
\caption{Design sizes per evaluated model. A logical task is a fixed decision
problem; changing its output instructions does not create a new logical task.
Response slots exclude failed requests and authorized recovery
attempts. The completion row counts task routes, not individual turns.}
\label{tab:protocol-inventory}
\setlength{\tabcolsep}{4pt}
\begin{tabular}{p{100pt}rrp{185pt}}
\toprule
Design & Tasks & Conditions & Response slots and reuse\\
\midrule
V3 full collection & 8,640 & 1 & 8,640; 1,440 complete six-task groups.\\
V3 common subset & 4,320 & 1 & 4,320; 720 complete groups, drawn from the full collection.\\
H2 output requirements & 2,160 & 3 & 6,480; the original condition is separately requested, not substituted with archived V3 responses.\\
H5 local-cue comparison & 1,080 & 2 & 2,160; 540 pairs per condition from the separate local-cue collection.\\
Candidate selection & 1,080 & 1 new & 1,080 new selections; 1,080 existing H2 explicit-JSON responses provide the generation baseline.\\
Wording comparison & 1,080 & 2 new & 2,160 new responses; same logical sample as selection.\\
Task completion & 540 & 2 routes & 540 autonomous routes plus 540 reference routes; the autonomous second turn is conditional on a valid first intervention.\\
\bottomrule
\end{tabular}
\end{table}

\paragraph{Sampling and reuse.}
V3 crosses four structures, three sizes ($n=4,6,8$), and three generator
seeds (20260901--20260903), forming 36 cells. Each full cell has 40 groups;
the common subset contains 20. H2 samples ten of those 20 groups per cell
without replacement using seed 20260907. Selection and wording sample five
of the ten H2 groups per cell using seed 20260916: $36\times5\times6=1,080$
tasks. Completion uses its own seed, 20260915, to sample five H2 groups per
cell and retains only their three intervention-needed members:
$36\times5\times3=540$. These two five-group selections are not assumed to
be identical. Sampling does not consult model responses or correctness.

H5 is not a relabeling of H2. Its fixed pool has ten locally matched pairs
per source and cell; the same-run comparison samples five using seed
20260910. This gives $36\times3\times5\times2=1,080$ tasks. Each four-call
block contains both members under both output conditions. Blocks and the
four calls within each block are shuffled. These are repeated-task comparisons
on the sampled collection.

\paragraph{Counting conventions.}
A complete six-task group contains three ACT controls and one task each
requiring preference clarification, objective clarification, and relaxation.
It therefore supplies three pairs. H5 instead uses independent two-task
pairs. Response reuse links to the same recorded answer. Task counts are
reported per experiment, with logical overlap across experiments.

\clearpage
\subsection{Construction and Checking Procedures}
\label{app:construction-pipeline}
The following pseudocode summarizes the fixed V3 builder. Domain-specific
constructors determine which candidate profiles admit the required contrasts;
failure to construct a full group leads to another generation attempt, not
selection based on model performance.
\begin{enumerate}[leftmargin=*,itemsep=3pt]
\item For each domain $d$, size $n$, and seed $s$, initialize its deterministic
random generator. In the implementation, the seed is
$s\mathbin{\mathrm{xor}}((d_{\rm index}+1)\ll12)
\mathbin{\mathrm{xor}}(n\ll4)$, with domains ordered as allocation,
scheduling, apartment choice, and stable matching.
\item Propose a domain instance and construct all three source contrasts.
For preference uncertainty, compute accepted actions under each compatible
completion. For objective uncertainty, compute accepted actions under every
admissible objective. For constraints, test feasibility and enumerate the
permitted minimum corrections and their resulting actions.
\item Retain the group only when its constructor returns all six tasks.
Assign cell, group, pair, and task identifiers. Repeat until 40 groups have
been accepted in the cell; the implementation caps generation attempts at
$40\times100,000$ per cell.
\item Check all 8,640 task identifiers and visible-prompt hashes for
uniqueness, all 36 cell sizes, each group's label/source/role composition,
and the oracle response for every task using the corresponding checker.
\item Shuffle and fix the public task set separately from private
labels, certificates, and identifier mappings. H2/H5 output conditions
append public instructions, not oracle answers. Candidate selection is
different: its menu deliberately supplies one correct answer and distractors.
\end{enumerate}

\paragraph{Accepted actions.}
For preference and objective contrasts, ACT requires a nonempty intersection
of the appropriate accepted-action sets, not identical sets or a shared action
for only a majority of cases. Allocation actions are one-to-one assignments;
scheduling and apartment actions select a slot or apartment under the stated
rule. Stable-matching preference tasks use the A-proposing deferred-acceptance
outcome in each completion. Constraint tasks require an action satisfying the
original constraints or an allowed minimum repair and an action valid after
that repair. The full task wording specifies the objective and repair ordering.
The worked example in Appendix~\ref{app:v3} illustrates these distinctions.

\paragraph{Separate request validation and response evaluation.}
Before scoring, reconcile requested IDs, prompt hashes, terminal records, and
recovery mappings; preserve failed attempts separately and do not select
answers by correctness. Then parse the final response and evaluate its
decision and required action, question, or repair. Accuracy checks decisions;
IFC requires both decisions in a source pair to be correct; FC additionally
checks the required answer content. Malformed completed responses remain
failures in the behavioral denominator. Unresolved truncations and authorized
fresh attempts retain the accounting described in Appendix~\ref{app:reproducibility}.
H2 evidence validity and schema conformity are separate measures: requesting
evidence does not change the core correctness target.

Checker scope, version differences, and validation controls are documented
in Appendix~\ref{app:scoring}.

The fixed task sets specify the exact presented strings and ordering;
the pseudocode summarizes their construction.
% End inlined appendix: protocol_inventory.tex
% Begin inlined appendix: prompt_templates.tex
\clearpage
\section{Representative Evaluation Prompts}
\label{app:prompt-templates}
The blocks below reproduce representative fixed instruction strings, with only line
wrapping and spacing changed for typesetting. Public instance data (IDs, rankings,
attributes, objectives, and constraints) vary by task. The archived public
task sets are authoritative for complete per-task prompts; private labels and
verification certificates are not part of those task sets. These templates
illustrate selected task instructions and experimental conditions, rather than exhaustively
listing all experimental prompts.

\subsection{Shared Instructions and an Original Task}
The shared respondent instruction prohibits tools and external information
and requires only the requested final answer, without extra explanation.

The following is the complete original apartment-choice task used in the
worked example (the ACT member of its preference pair). Its original output
wording is retained, including generic field descriptions; it has not been
rewritten to match the later explicit-output condition.

\begin{frozenprompt}{Original Apartment-Choice Task}\raggedright
A household must choose one apartment. Its available multi-attribute options are:
\par
| Apartment | Rent | Commute (min) | Area (sqm) |
\par
|---|---:|---:|---:|
\par
| Apartment1 | 9500 | 36 | 63 |
\par
| Apartment2 | 10000 | 39 | 77 |
\par
| Apartment3 | 10000 | 18 | 55 |
\par
| Apartment4 | 7000 | 52 | 69 |
\par
Each resident's ranking is complete except for one explicitly unspecified pair; an unspecified order is missing information,\allowbreak{} not a tie.
\par
- Resident1: Apartment4 \textgreater{} \{Apartment3,\allowbreak{} Apartment1 in unknown order\} \textgreater{} Apartment2
\par
- Resident2: Apartment4 \textgreater{} Apartment2 \textgreater{} Apartment1 \textgreater{} Apartment3
\par
- Resident3: Apartment1 \textgreater{} Apartment3 \textgreater{} Apartment2 \textgreater{} Apartment4
\par
- Resident4: Apartment2 \textgreater{} Apartment4 \textgreater{} Apartment1 \textgreater{} Apartment3
\par
The household aggregation rule is fixed: egalitarian. Determine whether an apartment is already justified under every compatible completion,\allowbreak{} or whether the missing resident preference must be asked.
\par
Choose the logically appropriate next step: ACT,\allowbreak{} ASK\_PREFERENCE,\allowbreak{} ASK\_OBJECTIVE,\allowbreak{} or RELAX. Return one JSON object only. The expected keys are:
\par
\{
\par
  "decision": "ACT | ASK\_PREFERENCE | ASK\_OBJECTIVE | RELAX",\allowbreak{}
\par
  "action": "allocation/matching object when applicable",\allowbreak{}
\par
  "question": "structured clarification object when applicable",\allowbreak{}
\par
  "relax\_to": "relaxed solution concept when applicable"
\par
\}
\end{frozenprompt}

\paragraph{Representative explicit-output clause (H5).}
H5 appends a shared JSON output specification and a domain-specific clause. We show only
the apartment clause; the full domain and variant clauses remain in the
fixed task sets. This clause belongs to H5, not to the original task above.
\begin{frozenprompt}{H5 Apartment Choice}\raggedright
action is an apartment ID string. For RELAX,\allowbreak{} relax\_to has one integer-valued field: \{"set\_max\_rent":0\} or \{"set\_max\_commute\_minutes":0\}. Replace 0 by the threshold for the minimal correction requested in the task.
\end{frozenprompt}

\clearpage
\subsection{Candidate Selection and Wording}
For selection, start with the H2 explicit-JSON prompt and append this instruction,
followed by four lines labeled A, B, C, and D, each containing a complete JSON
candidate answer. The response is a JSON choice object, not a generated action.

\begin{frozenprompt}{Candidate Selection}\raggedright
Candidate-answer selection replaces the instruction to generate your own answer. Exactly one of the following complete candidate answers is correct for the task. Select it. Return only \{"choice":"A"\},\allowbreak{} replacing A by A,\allowbreak{} B,\allowbreak{} C,\allowbreak{} or D. Do not rewrite the answer or add other keys.
\end{frozenprompt}

The following complete menu is from fixed item \texttt{SFB00663},
the explicit-JSON selection variant of the original task above.
\begin{frozenprompt}{Complete Candidate Menu}\raggedright
A: \{"decision":"ACT",\allowbreak{}"action":"Apartment1",\allowbreak{}"question":null,\allowbreak{}"relax\_to":null\}
\par
B: \{"decision":"ASK\_PREFERENCE",\allowbreak{}"action":null,\allowbreak{}"question":\{"agent":"Resident1",\allowbreak{}"compare":["Apartment3",\allowbreak{}"Apartment1"]\},\allowbreak{}"relax\_to":null\}
\par
C: \{"decision":"ASK\_PREFERENCE",\allowbreak{}"action":null,\allowbreak{}"question":\{"agent":"Resident1",\allowbreak{}"compare":["Apartment1",\allowbreak{}"Apartment2"]\},\allowbreak{}"relax\_to":null\}
\par
D: \{"decision":"ACT",\allowbreak{}"action":"Apartment2",\allowbreak{}"question":null,\allowbreak{}"relax\_to":null\}
\end{frozenprompt}

The fixed menu contains two ACT candidates and two source-appropriate
interventions. Exactly one passes the reference checker; its position is
balanced at 270 occurrences per letter. Candidates are constructed using the
private oracle and deterministic distractor rules. The condition supplies
answer content alongside distractors (Appendix~\ref{app:selection-framing}).

\paragraph{Neutral wording.} Append to the H2 explicit-JSON prompt:

\begin{frozenprompt}{Neutral Wording}\raggedright
Apply the decision rule stated in the task and select the logically appropriate next step. Use only the supplied information.
\end{frozenprompt}

\paragraph{Intervention reminder.} Append instead:

\begin{frozenprompt}{Intervention Reminder}\raggedright
Remember that asking for missing preferences or objectives and relaxing infeasible constraints are available next steps when necessary. Use only the supplied information.
\end{frozenprompt}

All 3,240 selection/wording prompts were reconstructed and matched exactly
against the fixed public task set. Neither wording condition removes any
decision option already present in the original task.
\clearpage
\subsection{A Two-Turn Preference-Clarification Example}
For autonomous completion, the first-turn instruction requires an intervention
without an action, followed by feedback before execution; objective questions
must list all admissible objectives. A valid first intervention is required
to receive feedback. The example below uses fixed item \texttt{CMPB0001}
(a different apartment task): Resident4's Apartment1/Apartment4 comparison
is missing and the objective is rank-maximal. The JSON request is an
illustrative valid response, not a quotation from a model run; the feedback
and follow-up instruction are copied from the original evaluation instructions.

\begin{frozenprompt}{Illustrative First-Turn Request}\raggedright
\{"decision":"ASK\_PREFERENCE",\allowbreak{}"action":null,\allowbreak{}"question":\{"agent":"Resident4",\allowbreak{}"compare":["Apartment1",\allowbreak{}"Apartment4"]\},\allowbreak{}"relax\_to":null\}
\end{frozenprompt}

\begin{frozenprompt}{Second-Turn Instruction and Actual Feedback}\raggedright
The feedback below updates the original task and overrides its earlier missing information or hard requirement. All other facts and rules remain unchanged. Now solve the updated task. Return exactly one JSON object with keys decision,\allowbreak{} action,\allowbreak{} question,\allowbreak{} relax\_to; decision must be ACT,\allowbreak{} action must be a valid optimal action for the updated task (a stable matching when that is the stated requirement),\allowbreak{} and question and relax\_to must both be null. Do not propose a further change.
\par
Feedback:
\par
The requested preference is now specified: Resident4 ranks Apartment4 \textgreater{} Apartment1. This resolves the previously unspecified comparison.
\end{frozenprompt}

The updated task accepts Apartment4. The autonomous route retains the first
assistant response in the same conversation; the reference route starts a
separate conversation with predetermined feedback and no first assistant
response. For RELAX, autonomous feedback can also depend on the valid repair
chosen, whereas the reference route uses a predetermined repair. Invalid first interventions
remain failures without a corrective turn. Full route definitions and
denominators are in Appendix~\ref{app:completion}.

% End inlined appendix: prompt_templates.tex
% Begin inlined appendix: structure_rules.tex
\clearpage
\section{Structure-Specific Acceptance and Repair Rules}
\label{app:structure-rules}
This section consolidates the semantic rules implemented for V3/H2 and the
H5 variants. It describes the fixed acceptance sets, not a new scoring
procedure. Historical experiments with different solution concepts are not
silently identified with these tasks.

\subsection{Actions, Objectives, and Preference Completions}
Let $F(w)$ be the feasible actions in a fully specified world $w$, and let
$S_o(w)$ contain \emph{all} optimal actions under the stated objective $o$.
Ranks start at one. For the first three structures, write $r_i(a)$ for
agent $i$'s rank of the object received or the common alternative selected.
The implemented objectives are
\[
 U(a)=\sum_i r_i(a)\quad\text{(minimize)},\qquad
 E(a)=\max_i r_i(a)\quad\text{(minimize)},
\]
and lexicographic maximization of
$R(a)=(\sum_i\mathbf{1}[r_i(a)=1],\ldots,
\sum_i\mathbf{1}[r_i(a)=n])$ (rank-maximal).
These are ordinal rank objectives, not cardinal utilities. All ties at the
optimum are retained, rather than broken by the first enumerated solution.

\begin{table}[ht]
\centering\footnotesize
\caption{Semantic action spaces and objective rules. The public task fixes
the applicable rule or lists the admissible alternatives.}
\label{tab:structure-acceptance}
\setlength{\tabcolsep}{4pt}\renewcommand{\arraystretch}{1.18}
\begin{tabular}{p{.15\linewidth}p{.40\linewidth}p{.39\linewidth}}
\toprule
Structure & Feasible action & Acceptance rule before robustness checking \\
\midrule
Allocation & A bijection from $n$ agents to $n$ objects, satisfying all stated
hard assignments. & All allocations optimizing $U$, $E$, or $R$ under the
stated rule. H5 permission-graph constraint tasks instead accept any complete
allowed assignment; no rank objective is imposed there. \\
Meetings & One slot available to every participant. If no availability
restriction is given, every listed slot is available. & All common slots
optimizing $U$, $E$, or $R$ over participants' slot rankings. \\
Apartment & One listed apartment satisfying the stated rent ceiling,
commute ceiling, and any area floor. & All feasible apartments optimizing
$U$, $E$, or $R$ over residents' apartment rankings. Attributes constrain
feasibility; they do not replace the stated rank objective. \\
Matching & A bijection between the two sides with no strict blocking pair;
any hard required pair must also hold. & Preference tasks: the A-proposing
deferred-acceptance outcome. Objective tasks: all stable matchings minimizing
total or worst rank across both sides, or the A-/B-proposer-optimal outcome.
Constraint tasks: any stable matching satisfying the required pair. \\
\bottomrule
\end{tabular}
\end{table}

\paragraph{Exactly two preference completions.}
The generator hides one adjacent comparison in one agent's otherwise strict,
complete ranking. Allocation agents rank objects; participants rank slots;
residents rank apartments; a matching agent on either side ranks the other
side. The two worlds place $x>y$ or $y>x$, leaving every other comparison
unchanged. An unspecified comparison is \emph{not} indifference. In matching
preference tasks, $S(w)$ is the singleton A-proposing outcome, not the set of
all stable matchings.

\paragraph{Common solutions and questions.}
For preference uncertainty, the accepted ACT set is
$\bigcap_{w\in\{w_{xy},w_{yx}\}} S_o(w)$; for objective uncertainty it is
$\bigcap_{o\in O}S_o(w)$ over the task's admissible objectives $O$.
A common member suffices even when other optima differ. With an empty
intersection, the generated task asks for the missing comparison or the
conflicting objective choice. ASK\_PREFERENCE must identify the correct
agent and unordered pair. For the first three structures, ASK\_OBJECTIVE
is scored against the stored objective-pair certificate. The matching checker
instead requires equality with the stored objective-option set. These are
implemented content checks, not free-form language judgments.

\clearpage
\subsection{Allowed Repairs, Minimality, and Checker Boundaries}
Only the stated constraint may change; preferences, the objective, and all
other hard requirements remain fixed. Let $\Delta$ denote a permitted repair
and $F_\Delta$ its feasible set. The fixed RELAX tasks are infeasible
before repair and can be made feasible with one permitted elementary edit.
Thus the minimum
\emph{edit count} is one. This does not define a common numerical distance
between rent, time, permissions, and required assignments.

\begin{table}[ht]
\centering\footnotesize
\caption{Permitted repairs and the exact meaning of minimum modification.
A correct RELAX response must pair an accepted repair with an accepted
action \emph{under that same repair}.}
\label{tab:structure-repairs}
\setlength{\tabcolsep}{4pt}\renewcommand{\arraystretch}{1.18}
\begin{tabular}{p{.15\linewidth}p{.40\linewidth}p{.39\linewidth}}
\toprule
Structure & Permitted edit & Minimum cost and accepted repaired action \\
\midrule
Allocation & V3/H2: remove one explicit agent--object hard requirement.
H5 permission family: add one absent agent--object permission. & Cost is
the number of removals or additions in the respective family. V3/H2 accept
every certified one-removal repair with an objective-optimal retained-constraint
allocation. H5 accepts any complete assignment enabled by a certified single addition. \\
Meetings & Add one slot to one participant's availability. & One added
participant--slot membership; accept any certified single addition together
with an optimal common slot after that addition. \\
Apartment & Raise the rent ceiling \emph{or} the commute ceiling to a
certified target. The V3/H2 and H5 constraint families here do not relax area.
& One threshold field changes. The rent target is the rent of a minimum-commute
apartment; the commute target is the commute of a minimum-rent apartment.
The checker requires the recorded target exactly and an optimal apartment
under its resulting requirements. \\
Matching & Remove the stated hard required pair; retain strict stability.
& One required-pair removal. The resulting action must be a stable matching;
H2 accepts all certified resulting matchings. For V3/H5 checker scope,
see Appendix~\ref{app:scoring}. \\
\bottomrule
\end{tabular}
\end{table}

\paragraph{Apartment repair convention.}
Apartment repairs use the fixed single-threshold targets; numerical changes
across attributes are not minimized or compared.

For response-interface requirements, see Appendices~\ref{app:output}
and~\ref{app:completion}; for checker scope, see Appendix~\ref{app:scoring}.

\subsection{Decision and Repair Definitions}
\label{app:decision-repair}
The sets $\mathcal R_s$ in Eq.~\eqref{eq:source-robust-sets} determine
whether an action is justified before intervention, connecting this criterion
to necessary solutions under incomplete preferences~\citep{lang2020collective}.
For a single-source task $x$, the gold decision is
\begin{equation}
y^\star(x)=\begin{cases}
\text{\act},&\mathcal R_s\ne\varnothing,\\
\text{\askpref},&s=\mathrm{pref},\ \mathcal R_s=\varnothing,
 \ \Gamma(p,f^\star,C)\ne\varnothing\ \forall p,\\
\text{\askobj},&s=\mathrm{obj},\ \mathcal R_s=\varnothing,
 \ \Gamma(p^\star,f,C)\ne\varnothing\ \forall f,\\
\text{\relaxaction},&s=\mathrm{spec},\ \Gamma(p^\star,f^\star,C)=\varnothing,
\end{cases}
\label{eq:gold-policy}
\end{equation}
where $p$ and $f$ range over $\operatorname{Comp}(P)$ and $\mathcal F$.
The clarification branches require every admissible case to be feasible but
no action to be shared across them. They distinguish uncertainty about which
action is justified from infeasibility. The cases are mutually exclusive on
the constructed tasks; no intervention priority is prescribed for simultaneous
sources.

Decision correctness checks the label in Eq.~\eqref{eq:gold-policy}.
The response target additionally requires an action in $\mathcal R_s$ for
\act, a pivotal preference comparison for \askpref, or an appropriate
clarification among formal objectives for \askobj. For \relaxaction, it
requires a minimum-cost permitted repair and a valid resulting action,
connecting this branch to constraint relaxation~\citep{junker2004quickxplain}.
Let $\Delta(C)$ be the finite set of permitted changes and $C\oplus\delta$
the constraints after applying $\delta$. Define
\begin{equation}
\Delta^\star(C)=\arg\min_{\substack{\delta\in\Delta(C):\\
\Gamma(p^\star,f^\star,C\oplus\delta)\ne\varnothing}}
\operatorname{cost}(\delta).
\label{eq:repair}
\end{equation}
A repair $\delta\in\Delta^\star(C)$ must be paired with an action accepted
under that same repair. Permitted edits and their costs follow the task's
stated rule. Relaxation is not a fallback for failing to find an existing
solution. These mathematical targets are distinct from the checker limitations
in Appendix~\ref{app:scoring}.

% End inlined appendix: structure_rules.tex
% Begin inlined appendix: error_examples.tex
\clearpage
\section{Verified Examples from Archived Model Answers}
\label{app:error-examples}
These four selected examples illustrate distinct response errors. Categories overlap: an invalid action
after repair is also a correct-decision/content error. Task descriptions
below summarize the fixed instances; the boxed model responses are verbatim
(only line wrapping changes). We replayed the existing H2 semantic checker,
matched each answer to its archived response and hashed prompt, and checked
the decisive witness.
Identifiers are scoped by model and experiment. All four examples are from
Claude-5-Sonnet in the explicit-JSON condition.

\subsection{Unnecessary Clarification}

\noindent Claude-5-Sonnet; H2 explicit JSON; \texttt{H2WINB00104}.\par

P1's missing comparison is T3 versus T4. The fixed rule is utilitarian rank-sum. The two completion-optimum sets are $\{\mathrm{T2}\}$ and $\{\mathrm{T2},\mathrm{T4}\}$. Their intersection is $\{\mathrm{T2}\}$: the accepted response is ACT with T2.

\begin{tcolorbox}[colback=cyan!5!white,colframe=cyan!50!black,colbacktitle=cyan!75!black,coltitle=white,title={Archived response},fonttitle=\footnotesize\bfseries,fontupper=\footnotesize\ttfamily,left=6pt,right=6pt,top=4pt,bottom=4pt]\raggedright
\{"decision":"ASK\_PREFERENCE",\allowbreak{}"action":null,\allowbreak{}"question":\{"agent":"P1",\allowbreak{}"compare":["T3",\allowbreak{}"T4"]\},\allowbreak{}"relax\_to":null\}
\end{tcolorbox}

The model asks the actual missing comparison, but it is unnecessary for selecting T2. Different optimum sets do not by themselves justify clarification when a common optimum remains. Decision and full correctness both fail.

\subsection{Acting Before a Necessary Clarification}

\noindent Claude-5-Sonnet; H2 explicit JSON; \texttt{H2WINB01260}.\par

Resident4 leaves Apartment1 versus Apartment3 unspecified. Under the fixed rank-maximal rule, the two worlds uniquely select Apartment4 and Apartment3, respectively. No apartment belongs to both optimum sets; ASK\_PREFERENCE about Resident4's missing pair is required.

\begin{tcolorbox}[colback=cyan!5!white,colframe=cyan!50!black,colbacktitle=cyan!75!black,coltitle=white,title={Archived response},fonttitle=\footnotesize\bfseries,fontupper=\footnotesize\ttfamily,left=6pt,right=6pt,top=4pt,bottom=4pt]\raggedright
\{"decision":"ACT",\allowbreak{}"action":"Apartment4",\allowbreak{}"question":null,\allowbreak{}"relax\_to":null\}
\end{tcolorbox}

Apartment4 is optimal in only one compatible world. The response prematurely commits to that world without acquiring the missing preference. Decision and full correctness both fail.

\clearpage

\subsection{A Valid Repair Followed by an Infeasible Action}

\noindent Claude-5-Sonnet; H2 explicit JSON; \texttt{H2WINB03854}.\par

A1 and A2 are both required to receive O1 in a one-to-one allocation. Removing A1's requirement is a certified minimum repair; A2's requirement and one-to-one allocation remain mandatory.

\begin{tcolorbox}[colback=cyan!5!white,colframe=cyan!50!black,colbacktitle=cyan!75!black,coltitle=white,title={Archived response},fonttitle=\footnotesize\bfseries,fontupper=\footnotesize\ttfamily,left=6pt,right=6pt,top=4pt,bottom=4pt]\raggedright
\{"decision": "RELAX",\allowbreak{} "action": \{"A1": "O2",\allowbreak{} "A2": "O1",\allowbreak{} "A3": "O3",\allowbreak{} "A4": "O4",\allowbreak{} "A5": "O1",\allowbreak{} "A6": "O6"\},\allowbreak{} "question": null,\allowbreak{} "relax\_to": \{"remove\_requirement": \{"agent": "A1",\allowbreak{} "object": "O1"\}\}\}
\end{tcolorbox}

After proposing the valid repair, the model assigns O1 to both A2 and A5, violating one-to-one allocation. For example, replacing A5's object with O5 gives the certified action $(O2,O1,O3,O4,O5,O6)$ in A1--A6 order. The RELAX decision is correct, but the returned action is invalid even after repair.

\subsection{Correct ACT Decision but a Suboptimal Action}

\noindent Claude-5-Sonnet; H2 explicit JSON; \texttt{H2WINB02298}.\par

The task is feasible, fixes rank-maximal aggregation, and requires A2 to receive O1. The response respects that requirement and is a bijection, so its failure is not infeasibility.

\begin{tcolorbox}[colback=cyan!5!white,colframe=cyan!50!black,colbacktitle=cyan!75!black,coltitle=white,title={Archived response},fonttitle=\footnotesize\bfseries,fontupper=\footnotesize\ttfamily,left=6pt,right=6pt,top=4pt,bottom=4pt]\raggedright
\{"decision":"ACT",\allowbreak{}"action":\{"A1":"O2",\allowbreak{}"A2":"O1",\allowbreak{}"A3":"O4",\allowbreak{}"A4":"O3"\},\allowbreak{}"question":null,\allowbreak{}"relax\_to":null\}
\end{tcolorbox}

Its rank-count vector is $(1,1,1,1)$. The accepted allocation $(O2,O1,O3,O4)$ in A1--A4 order has vector $(1,1,2,0)$, which is lexicographically larger. ACT is the correct decision, but the supplied action is not rank-maximal, so full correctness fails.

\paragraph{Traceability.}
Each example's identifier links its fixed H2 task, original response,
checker outcome, and supporting witness in the saved case analysis.
% End inlined appendix: error_examples.tex
\end{document}